# Precision in Rice Variety Classification using Stacking-Based Ensemble Learning

**Md. Masudul Islam** [1] (Corresponding Author)
Department of Computer Science and Engineering, Jahangirnagar University, Dhaka, Bangladesh
masudulislam11@gmail.com
**ORCID**: 0000-0001-7643-5420
**Address:** House-36, Road-18, Rupnagar Abashik, Pallabi, Dhaka-1216
**Contact No.:** +8801671118854

**Galib Muhammad Shahriar Himel** [2]
Department of Physics, Jahangirnagar University, Dhaka, Bangladesh
galib.muhammad.shahriar@gmail.com
ORCID: 0000-0002-2257-6751

**Md. Golam Moazzam** [3]
Department of Computer Science and Engineering, Jahangirnagar University, Dhaka, Bangladesh
khokan@juniv.edu

**Mohammad Shorif Uddin** [4]
Department of Computer Science and Engineering, Jahangirnagar University, Dhaka, Bangladesh
shorifuddin@juniv.edu
ORCID: 0000-0002-7184-2809

**Abstract**
Rice, a staple food for a significant portion of the global population, exhibits remarkable diversity in its varieties, presenting substantial challenges for accurate identification by consumers, traders, and farmers. This complexity often facilitates fraudulent practices, such as the unauthorized mixing of rice types, which undermines quality and trust in the supply chain. Despite its critical importance, existing research falls short of providing robust and efficient methods for precise rice variety classification based on external characteristics like color, size, and texture. To address this gap, our study introduces a comprehensive rice variety identification framework designed to enhance transparency and quality assurance. We developed a stacked ensemble model tailored for rice variety classification and curated a comprehensive dataset comprising 20 rice varieties, each distinguished by unique visual attributes. The proposed approach achieved an unprecedented classification accuracy of 100%. Furthermore, we integrated our model into a mobile application, enabling even novice users to effortlessly identify rice varieties using grain images from a smartphone camera. These findings underscore the transformative potential of advanced machine learning techniques in mitigating fraudulent practices and ensuring stringent rice quality control. Our work holds significant implications for agricultural stakeholders, paving the way for automated crop identification systems and advancing precision agriculture practices.

## 1. Introduction

Computer vision, a branch of artificial intelligence, enables computers to interpret visual data by analyzing digital images from cameras, high-resolution videos, and deep-learning models. This technological capability allows for precise identification and localization of objects, empowering computers to respond effectively to visual stimuli. It plays a pivotal role in tasks such as data classification. Advances in computer vision and artificial intelligence have catalyzed innovative research, particularly in automated rice variety identification. Rice, a staple food for approximately 80% of the population in Southeast Asia, holds significant cultural and economic importance as one of the first cultivated crops and a widely consumed commodity. Asia alone accounts for 90% of global rice cultivation, with rice serving as a staple for more than 60% of the world's population, emphasizing the growing need for automated variety identification methods in rice production (Bandumula, 2017). Globally, rice is cultivated in about 40,000 varieties, encompassing diverse categories such as short, medium, and long grains and brown, milled, and parboiled rice (Chauhan, 2017). According to FAOSTAT, global rice production reached 776.5 million metric tons in 2022 (FAO, 2023). Asian rice, scientifically known as Oryza sativa, includes numerous ecotypes or cultivars adapted to a wide range of environmental conditions and is grown on every continent except Antarctica. Notably, Bangladesh has emerged as a significant contributor to research on the genetic diversity and cultivation methods of rice varieties. Efforts to enhance rice crops have spanned over a century, driven by the diverse shapes, sizes, textures, and colors of rice grain kernels (Wu et al., 2012). Advances in computer vision have enabled the detection of diseases or damage in rice seeds through processes such as image acquisition, preprocessing, segmentation, and interpretation. These techniques facilitate seed classification and quantification. However, both consumers and farmers often encounter difficulties in identifying rice varieties and assessing quality, leaving them vulnerable to fraud from dishonest sellers. To address these issues, AI-based expert systems have been developed to identify rice varieties. The automation of rice variety identification presents notable challenges (Komal et al., 2023), largely due to the extensive diversity of strains cultivated worldwide. Each geographical region introduces unique complexities, with inherent impurities in rice grains further complicating the identification process. Research on automating rice variety identification remains limited. Traditional methods used to evaluate rice characteristics, essential for ensuring quality and variety, are not only time-consuming but also expensive and reliant on potentially harmful reagents. There is a critical need to transition from these conventional approaches to faster, automated methods for determining rice varieties. Such innovative strategies have the potential to revolutionize rice quality assessment by making it more efficient, cost-effective, and environmentally sustainable. This study aims to enhance rice variety classification using ensemble learning on a self-curated comprehensive rice grain image dataset. The major contributions of this study are:

i. Developed a comprehensive and well-structured dataset consisting of 20 distinct rice varieties.
ii. Extracted diverse deep features from the image dataset using hyperparameter-optimized deep learning models.
iii. Applied a stacking-based ensemble approach to the best-performing baseline models to improve classification accuracy.

iv. Developed a mobile application by integrating our machine learning model to identify rice varieties from grain images using a smartphone.

This article is organized as follows: in the first section, an introductory passage is presented. In the second section, a summarized literature review highlighting the pros and cons of related studies is provided. The third section presents a graphical depiction of the system architecture. In the fourth section, we describe the dataset, proposed methodology, and evaluation metrics. The fifth section discusses the experimental results, model deployment in the app, and various related aspects. In the sixth section, a comparative study is performed. Finally, the last section concludes the study and outlines future directions.

## 2. Literature Review

This section provides an overview of previous research on the detection and classification of rice varieties, followed by an overview of the primary contributions of this article. Classifying various types of food grains poses significant challenges, which require continuous efforts to improve classification accuracy. Marini et al. (Marini et al., 2002) used a counter-propagation artificial neural network to predict 11 rice varieties with nearly 90% precision. Hobson et al. (Hobson et al., 2007) proposed a digital image analysis technique for eight rice varieties, allowing the quantitative characterization of rice grains. The advancement continued in 2008 with Guzman and Peralta (Guzman and Peralta, 2008), who developed an artificial neural network that achieved close to 97% accuracy in classifying five varietal groups of rice in the Philippines. OuYang et al. (OuYang et al., 2011) introduced a neural network classifier based on color and texture characteristics, achieving 96.67% accuracy for five types of rice. Nagendra et al. (Nagendra et al., 2020) employed chemical tests for rice classification with an accuracy of 85%. Huang et al. (Huang et al., 2016) utilized Support Vector Machine techniques to classify 79 rice varieties with 79.74% accuracy, while Aki et al. (Aki et al., 2016) achieved 90.5% accuracy using the Non-Nested Generalization Algorithm on four rice types. Sumaryanti et al. (Sumaryanti et al., 2015) applied the learning vector quantization neural network Algorithm to classify ten rice grain varieties with 96.6% accuracy. The year 2016 saw the development of an efficient rice grain classification technique using backpropagation neural networks and wavelet decomposition (Singh et al., 2016). Kuo et al. (Kuo et al., 2016) applied the SRC Classifier to 30 rice varieties with an accuracy rate of 89.1%. Zareiforoush et al. (Zareiforoush et al, 2016) achieved an impressive 98.72% accuracy using an artificial neural network on a small rice dataset. In 2018, Lin et al. (Lin et al., 2018) introduced a deep convolutional neural network, achieving 95.5% accuracy in classifying three rice varieties, while Rexce J et al. (Rexce J et al., 2017) achieved 92.31% accuracy using the MLP neural network on 13 rice types. Wah et al. (Wah et al., 2018) classified three types of rice using image segmentation and the K-Nearest Neighbor classifier. Moving to 2019, Cinar and Koklu (Cinar and Koklu, 2019) applied various artificial intelligence methods to classify two types of rice with 93.02% accuracy. Hue et al. (Hue et al., 2019) utilized artificial neural networks to extract morphological features, achieving 95.5% accuracy. Dheer and Singh (Dheer and Singh, 2019) achieved 99.16% accuracy using machine learning classifiers and physical feature extraction on eight types of rice. Kumar and Javeed (. Kumar and Javeed, 2019) employed Multiclass SVM for classification with 92% accuracy. In 2020, Ibrahim et al. (Ibrahim et al., 2020) conducted a comparative analysis of rice classification using artificial neural networks, achieving 93.34% accuracy. Weng et al. (Weng et al., 2020) utilized hyperspectral imaging and deep learning techniques to classify rice varieties, achieving 98.57% accuracy. In 2022, Saxena et al. (Saxena et al., 2022) used popular machine learning methods, achieving an accuracy of 99.85% in five types of rice. Lee and Tay (Lee and Tay, 2022) applied convolutional neural networks to a small rice image dataset containing seven rice types. Jeyaraj et al. (Jeyaraj et al., 2022) proposed a deep learning technique for classifying five rice varieties, along with a computer-aided real-time variety detection system. In 2023, Tasci et al. (Tasci et al., 2023) used the LeeNet-5-based quantized neural network method to classify five rice types, achieving 99.87% accuracy. Iqbal et al. (Iqbal et al., 2023) developed a lightweight MobileNetV2 model to classify five types of rice with 99.7% accuracy. In 2024, Kang et al. (Kang et al., 2024) used fluorescence hyperspectral image data from 5 rice varieties and analyzed the principal components of the dataset. They achieved 95.3% accuracy using back-propagation neural networks and Random Forest. Another state-of-the-art study (Setiawan et al., 2024) applied the ensemble method using an SVM classifier on 3 rice varieties and achieved 96% accuracy. A recent study (Çifci and Kırbaş, 2024) presents a case study on classifying Cammeo and Osmancik rice species using the fusion of machine learning and explainable AI method.

**Table 1** represents a summarized view of various studies regarding rice variety classification.

**Table 1.** Summary Description of Related Studies

| Dataset Short Description | Data Size | Varie-ties | Classification Methods | Extracted Features |
|---|---|---|---|---|
| Italian rice is characterized by physical measurements (Marini et al., 2002) | 1779 | 11 | Counter-Propagation Artificial Neural Network (CP-ANN) | Physical measurements (e.g., size, shape) |
| Images of eight common rice varieties from the UK (Hobson et al., 2007) | 400 | 8 | Unsupervised Clustering Techniques | Grain length, aspect ratio, compactness, texture |
| Images of 52 rice grain samples from various agroecological zones (Guzman and Peralta, 2008) | 5720 | 52 | Multi-Layer Perceptron | 13 morphological features |
| Images of five rice varieties characterized by color and texture (Mousavi et al., 2011) | 500 | 5 | Segmentation And Back Propagation Neural Network-Based Classifier | 60 color and texture features |
| Bulk rice samples (Fajr, Hashemi, Daneboland, Gerde, Basmati, Domsiah, and Tailandi) with various morphological features (MousaviRad et al., 2012) | 2100 | 7 | Imperialist competition algorithm and SVM | Color, shape, texture attributes |
| Images of various Iranian rice kernels characterized by morphological features (Mousavi et al., 2012) | 1500 | 5 | Segmentation and Backpropagation neural network-based classifier | 18 morphological features |
| Indian Basmati Rice kernel samples analyzed for quality grading (Kaur et al., 2024) | Not Speci fied | 1 | Multi-Class SVM | 10 geometric features |
| Images of Basmati rice varieties captured under controlled conditions (Kambo et al., 2014) | 477 | 3 | Principal Component Analysis | Features derived from image processing and PCA |
| 8-bit Gray-scale image of Rice grains featured by Binary, Histogram, and Texture (Qadri et al., 2021) | 10800 | 6 | Logistic Model Tree, Regression, J48 Tree, Meta Bagging, Meta Attribute Selected Classifier | 43 features |
| 5 types of Indian Basmati rice image taken for MATLAB processing (Kaur et al., 2015) | 2917 | 5 | Average of Major axis length | 7 geometry features are extracted |
| Images of bulk rice grains characterized by color, texture, and wavelet features (Singh et al., 2016) | 1200 | 4 | BPNN, SVM, KNN, Naive Bayes | 18 color features, 27 texture features, 24 wavelet features |
| Diverse images of Japanese rice grain species (Lin et al., 2018) | 7399 | 3 | DCNN | Spectral features, morphological features, texture |
| Images of Indian rice (Basmathi, Idly, Samba, Ponni, and Ponni) grains in scattered and heap arrangements (Bhat et al., 2017) | Not Speci fied | 5 | Scattered and Heap Fashion Image Processing | Geometric features estimated from images |
| Images of Paw-San rice kernels classified based on quality grades (Wah et al., 2018) | Not Speci fied | 1 | KNN | Broken rice percentage |
| Turkish two types of rice grains to obtain morphological features (Cinar and Koklu, 2019) | 3810 | 2 | LR, MLP, SVM, DT, RF, NB and k-NN | Extracted 7 morphological features |
| Images of multiple Malaysian rice varieties categorized under controlled conditions (Ibrahim et al., 2020) | 90 | 3 | SVM and ANN | Morphological and color features |
| Hyperspectral images of China rice varieties (Weng et al., 2020) | 4320 | 10 | principal component analysis network | Spectral reflectance, color, morphology, texture |
| 75000 images of five Turkish rice varieties (Koklu et al., 2021) | 75000 | 5 | ANN, DNN, CNN | 24 color features, 11 morphological features, and 4 shape factors |
| 75000 Turkish rice grain images capturing physical characteristics (Cinar and Koklu, 2021) | 75000 | 5 | Not Specified | 106 features (12 morphological, 4 shape, 90 color) |
| Grayscale images of Vietnamese rice grains (Jeyaraj et al., 2022) | 857 | 9 | CNN | Various attributes extracted for classification |
| Malaysian rice images capturing various varieties (Lee and Tay, 2022) | 700 | 7 | ANN and CNN | Various image features |
| 75000 rice grain images collected from grocery stores (Saxena et al., 2022) | 75000 | 5 | Random Forest | Various attributes for quality control |
| 75000 sample images of Turkish rice varieties Iqbal et al., 2023) | 75000 | 5 | CNN | Various extracted features from images |
| 45000 images of Arborio, Basmati, and Jasmine rice varieties (Setiawan et al., 2024) | 45000 | 3 | Ensemble Learning | Image features processed with Bagging and SVM |
| Fluorescence hyperspectral data from multiple rice varieties (Kang et al., 2024) | 550 | 5 | Principal Component Analysis, BP Neural Network, Random Forest | Spectral features across different wavelengths |

The mentioned studies represent some contributions to rice research, yet many encounters significant issues. Challenges include small sample sizes and limited rice variety coverage, undermining the deep learning’s data-dependent nature. Additionally, new deep-learning methodologies are underutilized in rice studies. Addressing these gaps, our research employs an extensive dataset and advanced deep-learning techniques to improve rice variety classification.

### 3. System Architecture

Our proposed smartphone-based rice variety identification expert system is represented in **Figure 1**. The process is simple, i.e., any person can capture a rice grain image by zooming at least 5x on his smartphone camera through our Android app. Our app will automatically process the grain image in the back-end system where our developed model is installed. The system will predict the name of the rice variety and provide the result on the interface screen.

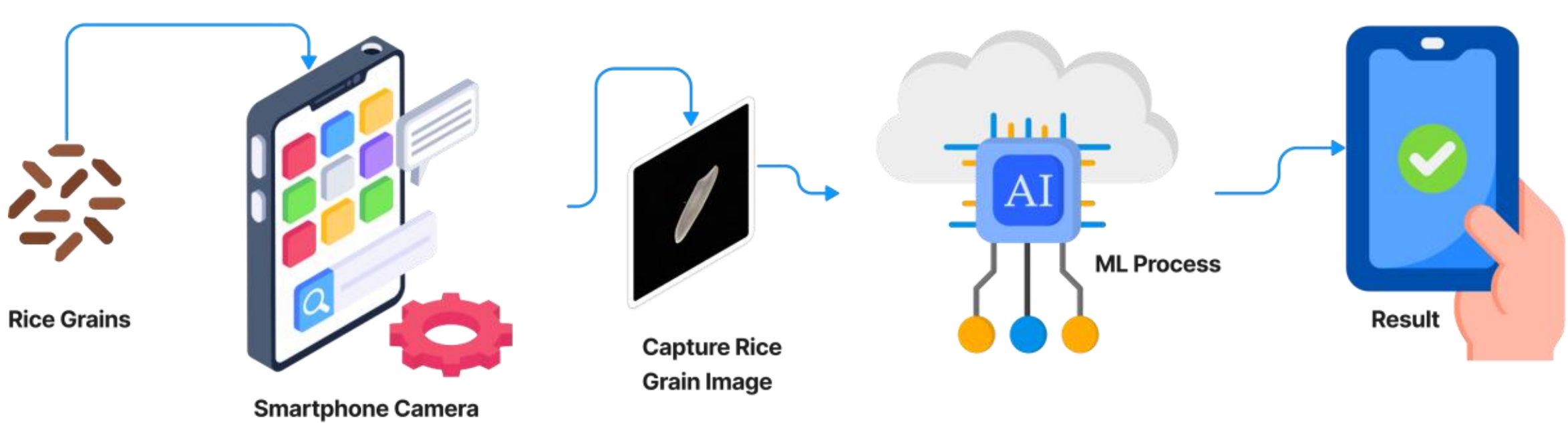

**Figure 1.** System Architecture of our Proposed Model

### 4. Methodology

Our proposed framework consists of several key stages: (i) developing a comprehensive and diverse dataset with unique visual attributes, (ii) applying various transfer learning models by tuning hyperparameters and extracting deep features, (iii) selecting the best-performing models as baselines and implementing stacking-based ensemble learning, (iv) mitigating potential overfitting and performing cross-validation, and (v) developing an Android application for classifying rice varieties using a smartphone camera.

Each of these components will be described in detail in the following subsections.

#### 4.1. Datasets

Obtaining clear images of rice grains is essential for accurately identifying different rice varieties. Since rice cultivars exhibit diverse distinguishing characteristics, such as morphology, shape, color, and texture, compiling a comprehensive rice dataset poses a significant challenge. In this study, we utilized two datasets for distinct purposes. The first is our self-curated dataset, which is publicly accessible in the 'Mendeley Data' repository (Islam et al., 2024) and comprehensively described in data article (Md. Masudul et al., 2024). The second is a well-known, publicly available rice dataset curated by Cinar and Koklu (Koklu and Cinar, 2021). In our proposed method, we have used our self-curated dataset which contains 20 rice varieties based on color and size criteria, as authenticated by agricultural experts. The selected varieties include Subol Lota, Bashmoti (Deshi), Ganjiya, Shampakatari, Sugandhi Katarivog, BR-28, BR-29, Paijam, Bashful, Lal Aush, BR-Jirashail, Gutisharna, Birui, Najirshail, Red Cargo, Polao (Katari), Polao (Chinigura), Amon, Shorna-5, and Lal Binni. We meticulously captured individual images of each grain from every variety at 5x magnification, resulting in approximately 4,500 images (225 samples per class). Using data augmentation techniques, we expanded the dataset to 27,000 images (1,350 samples per class). The augmentation methods employed—horizontal and vertical flipping, as well as rotations of 45°, 90°, and 180°—were carefully chosen to preserve the grains' key characteristics. Sample images from the dataset are presented in **Figure 2**. The dataset was split into training and testing sets at an 80:20 ratio, with the distribution detailed in **Table 2**.

Additionally, we utilized a second dataset for external validation to assess the model's potential in real-world applications, particularly when tested on previously unseen varieties and conditions. This dataset originally comprised 75,000 rice grain images spanning five varieties (15,000 images per class). From this, we carefully selected a subset of 3,500 images (700 images per class) to evaluate the generalizability of our proposed model.

**Table 2.** First Dataset Description

| Dataset Splitting | Images Numbers | Images Per Class |
|---|---|---|
| Training Set | 22,500 | 1,125 |
| Test Set | 4500 | 225 |
| Total | 27,000 | |

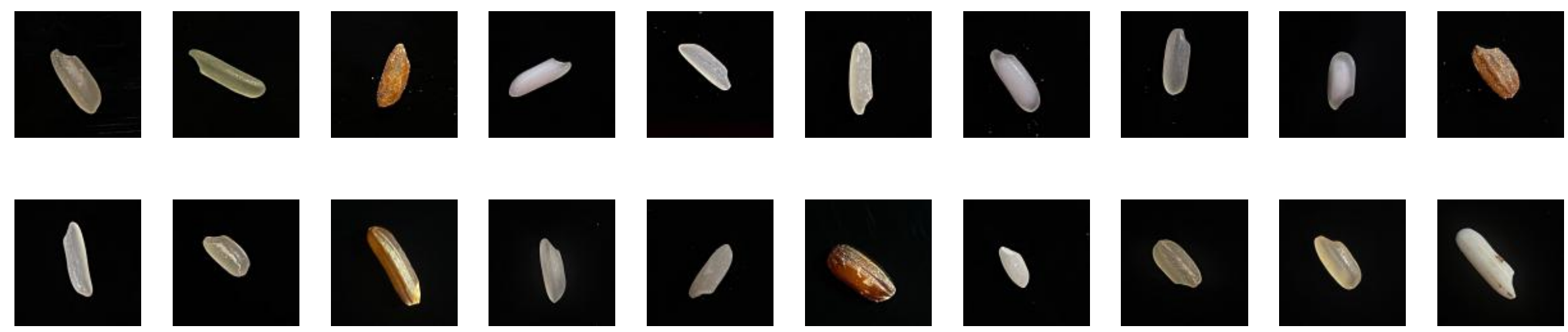

**Figure 2.** Rice Dataset Example

It is to be mentioned that, while the curated dataset includes diverse rice varieties captured under controlled conditions, it does not fully replicate real-world complexities such as mixed-grain samples or environmental variability. Future work will focus on expanding the dataset to include images from non-controlled environments and additional rice varieties sourced globally.

### 4.2. Proposed System Architecture

In this section, we present a comprehensive architectural framework for our experiments. The proposed approach consists of two main steps: first, extracting deep features of rice varieties using a variety of deep learning models, and second, improving classification accuracy by applying ensemble techniques to the best-performing models. For the experiment, we systematically selected models known for their exceptional performance across diverse classification tasks. **Figure 3** illustrates the architecture of our proposed ensemble model.

In the first stage, referred to as "Feature Extracting" the process begins with a dataset of rice images. Various transfer learning models are utilized to extract meaningful features from these images. These models learn patterns specific to the rice varieties, and the extracted features are grouped into multiple feature sets, such as Feature Set 01, Feature Set 02, and so on. Each feature set represents the unique characteristics identified by a specific transfer learning model. This stage lays the foundation for accurate classification by ensuring diverse and rich feature representations. In the second stage, titled "Ensemble: Stacking Method," the extracted feature sets are combined using a stacking-based ensemble approach. Fourteen meta-learners are employed to integrate these feature sets, leveraging the strengths of multiple models to improve overall predictive performance. Meta-learners are models designed to aggregate predictions from multiple base models, optimizing the overall performance by learning from their outputs. In the context of this study, meta-learners play a crucial role in the stacking-based ensemble approach, combining the strengths of individual classifiers to improve classification accuracy and generalizability. Among these 14 meta-learners, XGBoost is selected as the final model for classification. It uses the aggregated outputs from the stacking method to predict the rice variety. The output of this stage is the identified rice variety, which is displayed as the final result of the workflow.

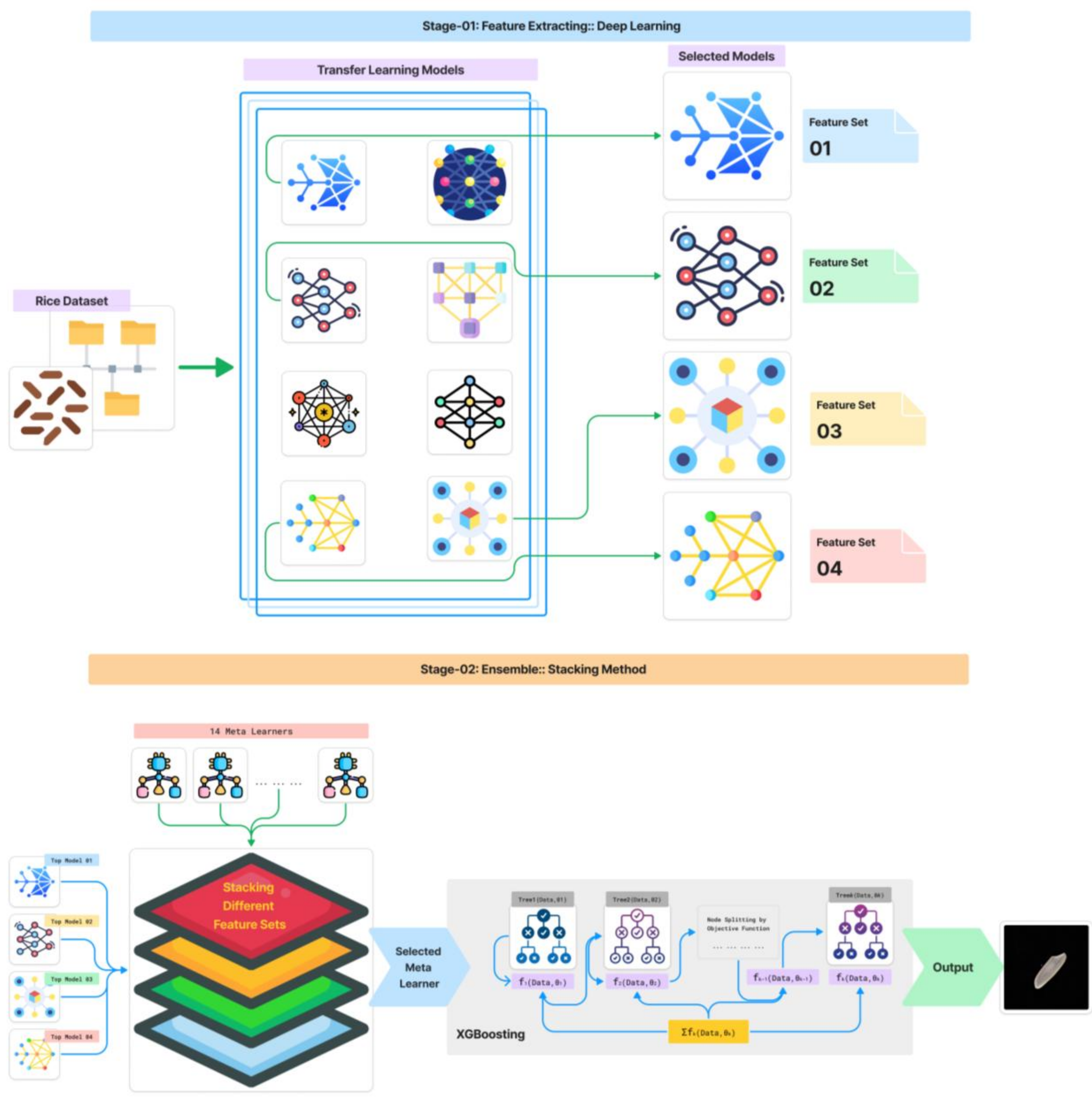

**Figure 3.** Proposed Ensemble Model Architecture

We individually applied ten deep learning models—VGG16, MobileNetV2, EfficientNetV2L, DenseNet201, NASNetLarge, ResNet50V2, Xception, InceptionV3, VGG19, and ResNet152V2—by performing hyperparameter tuning. Through this process, along with the integration of regularization techniques and dropout layers, we significantly improved the precision rate and reduced the likelihood of overfitting. Furthermore, we enhance accuracy through a stacking-based ensemble approach, amalgamating the predictions of the top 4 models out of 10 models with the aid of 14 meta-learners: Logistic regression, K-Nearest Neighbors, Support Vector Classifier, Decision Tree, Random Forest, AdaBoost Classifier, eXtreme Gradient Boosting Classifier, Gradient Boosting Classifier, Gaussian Nave Bayes, Passive Aggressive Classifier, Ridge Classifier, Stochastic Gradient Descent Classifier, Extra Trees Classifier, Multi-Layer Perceptron Classifier. Among these, XGBoost outperforms others. **Figure 4** shows the best-performing models' architecture.

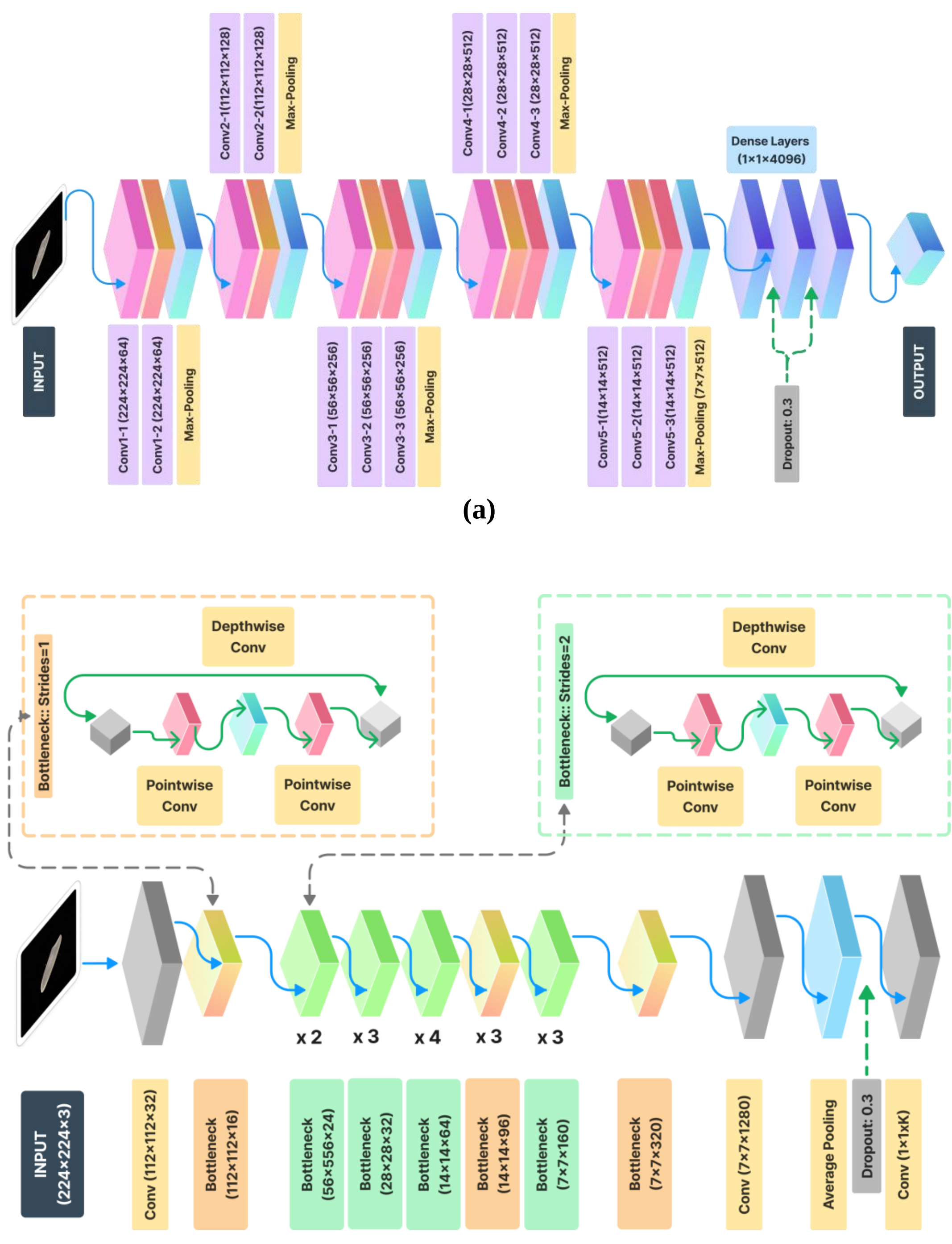

**(b)**

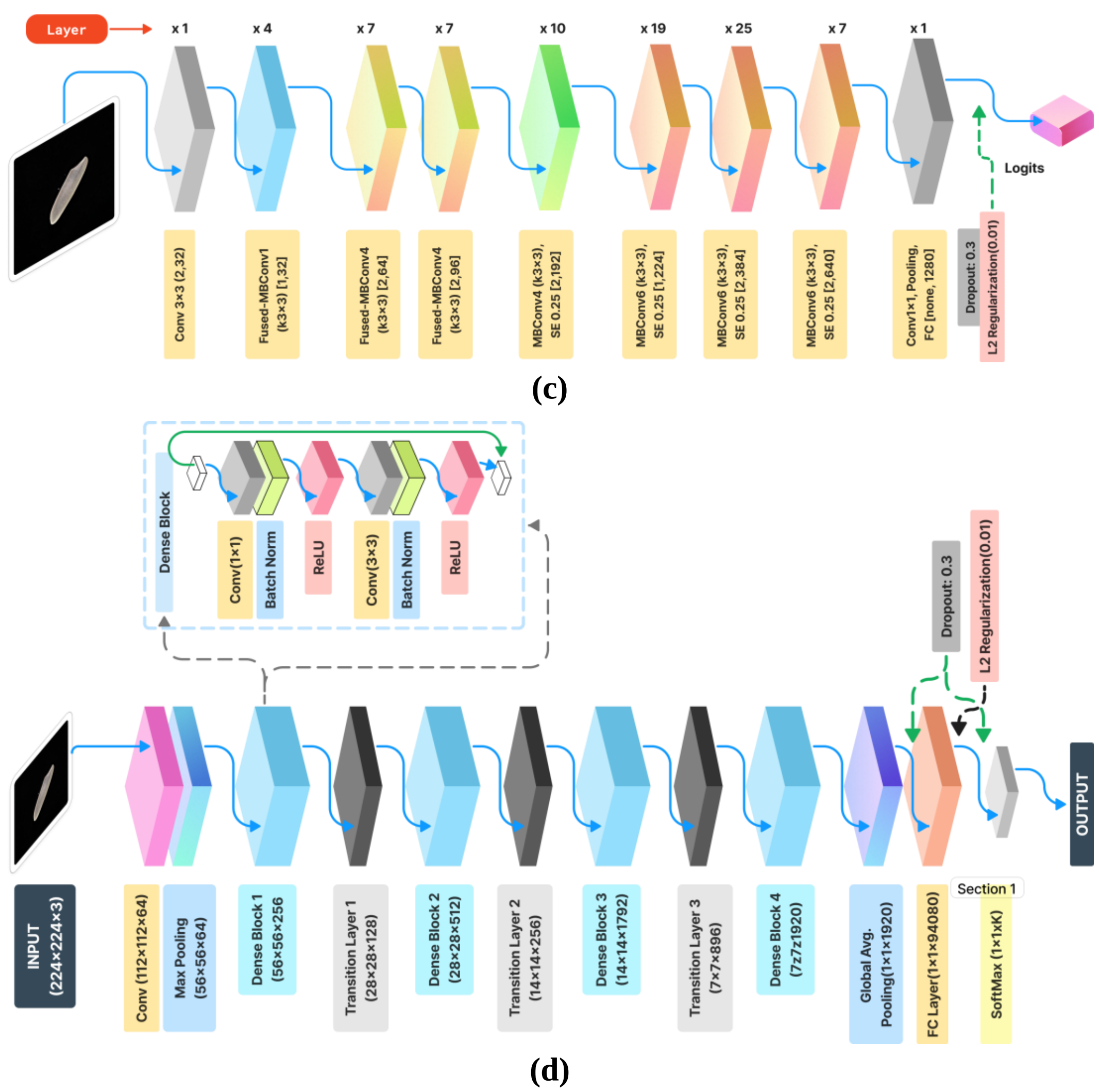

**(c)**

**(d)**

**Figure 4.** (a) VGG16 (b) MobileNetV2 (c) EfficientNetV2L (d) DenseNet201 architecture

## 4.3. Experimental Settings

Transfer learning leverages pre-trained convolutional neural network (CNN) models as foundational frameworks for new tasks, such as image classification. These pre-trained models are typically trained on the 'ImageNet' dataset, enabling them to learn fundamental image features like edges, lines, and textures, which are universally applicable to a wide range of tasks. By fine-tuning the hyperparameters of these models with a new dataset, they adapt to recognize more domain-specific features (Weerts et al., 2020), thereby improving classification accuracy, even with limited training data. Since most rice grain images appear quite similar to one another, it is challenging for even complex models to generalize and identify the most distinguishing features across different varieties. This similarity increases the risk of overfitting, which we also addressed using regularization and dropout techniques.

The details of the hyperparameter configurations are presented in **Table 3**. Before starting the primary experiment, we conducted an initial investigation to determine the optimal configuration for key training parameters such as the number of epochs, batch size, and learning rate. This preliminary exploration was essential to fine-tune our model training process and ensure its effectiveness. By systematically varying these parameters and evaluating their impact on the performance of the model, we aimed to identify the most suitable settings that would facilitate efficient convergence and maximize the predictive accuracy of the model. This meticulous pre-experimentation allowed us to establish a robust foundation for our subsequent experimental endeavors and optimize the overall training process for optimal outcomes.

We took the Adam optimizer, as it combines the benefits of two other extensions of stochastic gradient descent, namely, AdaGrad and RMSProp. The Adam optimizer computes individual adaptive learning rates for different parameters from estimates of the first and second moments of the gradients.

In the majority of cases, our model exhibited its highest level of performance, characterized by significant improvements in accuracy and convergence, by the time it completed the 20$^{th}$ epoch of training. As a result of observing this trend across multiple experiments, we made the strategic decision to extend the training process slightly beyond this point, allowing the model to undergo additional epochs of learning. By increasing the number of epochs to 25, we aimed to provide the model with ample opportunity to further refine its internal representations and fine-tune its parameters. This extension of the training duration beyond the point of optimal performance was driven by our desire to maximize the model's potential and ensure that it had fully converted to a stable and highly accurate state. After conducting several experiments with a wide range of batch sizes, including 8, 16, 32, 64, 128, and 256 we observed that a batch size of 32 consistently produced the most promising outcomes. This finding underscores the importance of batch size selection in neural network training, as it directly impacts both training efficiency and model performance. While larger batch sizes indeed accelerate the training process by processing more samples per iteration, they often lead to suboptimal accuracy and generalization due to limited exploration of the parameter space. In contrast, in our cases, smaller batch sizes tend to yield lower error rates in the validation set, indicating enhanced model precision. Therefore, the choice of batch size involves a careful balance between computational efficiency and model efficacy, with batch size 32 emerging as a favorable compromise in our experiments, striking a balance between training speed and predictive accuracy. When considering the learning rate, we experimented with various values to optimize model training. Initially, a learning rate of 0.001 yielded inconsistent results, indicating that the rate was too high for stable convergence. Subsequently, we systematically reduced the learning rate and observed notable improvements in the consistency of the results as we approached values below 0.0001. However, further reductions below this threshold did not yield significant enhancements in model performance. After careful evaluation, we selected a learning rate of 0.0001 as the optimal compromise between stability and duration of training. This choice allowed for sufficient model convergence while mitigating the risk of divergence or slow convergence associated with higher or excessively low learning rates, respectively. By prominent this balance, we ensured effective training dynamics and facilitated the model's ability to learn meaningful patterns from the data, ultimately enhancing its predictive accuracy and generalization capabilities.

To mitigate overfitting and enhance generalization in our deep learning models, we applied dropout regularization and L2 weight regularization across the four selected base models: EfficientNetV2L, VGG16, MobileNetV2, and DenseNet201. For each model, a dropout rate of 0.3 was applied after the fully connected layers, ensuring a balance between regularization and effective learning. Additionally, L2 regularization (also known as weight decay) with a penalty coefficient of 0.001 was introduced in the final dense layers to penalize large weight magnitudes, thereby constraining the complexity of the learned models and reducing the risk of overfitting.

In EfficientNetV2L, dropout was applied to the penultimate layer, while the final classification layer used L2 regularization to optimize weight distribution. In VGG16, dropout rates of 0.3 were added after both the first and second fully connected layers to ensure regularized feature extraction. MobileNetV2, known for its efficient architecture, also benefited from dropout in its final pooling layer, maintaining a dropout rate of 0.2 to preserve its lightweight nature. DenseNet201, characterized by its densely connected layers, incorporated a dropout rate of 0.4 due to its higher parameter count, ensuring effective regularization throughout its complex network structure. By integrating these techniques, we observed a notable reduction in validation loss across all models, with validation accuracy consistently stabilizing and achieving superior generalization performance compared to models without these regularization techniques.

**Table 3.** Hyperparameter Settings

| Hyperparameter | Optimization Space |
|---|---|
| Epoch | *25* |
| Batch size | *32* |
| Learning rate | *0.0001* |
| Optimizer | *Adam* |
| Loss function | *Categorical cross-entropy* |
| Activation Function | *softmax* |
| Class mode | *categorical* |

### 4.4. Stacking-based Ensemble Approach

To enhance our classification accuracy, we have applied the stack ensemble learning technique that combines the predictions of multiple base models to improve overall predictive performance. The process involves training multiple base learners. $(\{M_1, M_2, \dots, M_k\})$, where each model $(M_i)$ is trained independently on the same dataset or subsets of it. These models generate predictions. $(\widehat{y_1}, \widehat{y_2}, \dots, \widehat{y_k})$ , Which serve as inputs to a higher-level meta-model $(M_{\text{meta}})$. The meta-model learns to optimally combine these predictions to produce the final output. $(\widehat{y_{\text{final}}})$. The base model outputs $(h(X) = [\widehat{y_1}, \widehat{y_2}, \dots, \widehat{y_k}])$, The meta-model can be expressed as $(\widehat{y_{\text{final}}} = M_{\text{meta}}(h(X)))$. The training of the meta-model is done using cross-validation to prevent overfitting and ensure robust generalization. This stack ensemble is particularly effective in leveraging the strengths of diverse models while mitigating individual weaknesses.

### 4.5. Evaluation Metrics

To evaluate our baseline models and the final ensemble learning, we have relied on a confusion matrix, which contains four basic metrics: precision, sensitivity, F-1 score, and accuracy. These metrics provide insight into the model's ability to correctly classify rice grains into their respective varieties.

***Precision***: It indicates how accurately the model identifies rice grains belonging to a specific variety. Mathematically, precision is calculated as follows:

$$Precision = \frac{\text{True Positives}}{\text{True Positives} + \text{False Positives}}$$

***Recall***: Recall, also known as sensitivity or true positive rate, represents the model's ability to identify all instances of a particular rice variety correctly. Mathematically, recall is calculated as follows:

$$\text{Recall} = \frac{\text{True Positives}}{\text{True Positives} + \text{False Negatives}}$$

***F-1 score***: The F-1 score reaches its best value at 1 (perfect precision and recall) and worst at 0. Mathematically, the F-1 score is calculated as follows.

$$F1 - score = 2 \times \frac{\text{Precision} \times \text{Recall}}{\text{Precision} + \text{Recall}}$$

***Accuracy***: It represents the model's ability to correctly classify rice grains into their respective varieties. Mathematically, accuracy is calculated as follows:

$$Accuracy = \frac{\text{True Positives} + \text{True Negatives}}{\text{Total Number of Instances}}$$

We have also used AUC-ROC for our multiclass classification using the 'OvO' (One vs One) approach (Galar et al., 2011). For $k$ classes of rice, this results in $\frac{k \times (k-1)}{2}$ binary classifiers. ROC curves can be generated for each binary classifier, where the 'True Positive Rate' and 'False Positive Rate' are calculated based on the classification results between the two classes. The overall multiclass ROC curve can then be constructed by aggregating the TPR and FPR across all binary classifiers. Also, to cross-check our achieved validation accuracy our proposed ensemble model is trained *m* times, each time using *m-1* subsets as training data and one subset as validation data. This process allows a thorough assessment of the performance of the model in multiple iterations, helping to detect overfitting and produce more reliable performance estimates. Since our achievement is 100% test accuracy, we need proper cross-validation for our result. Finally, we have tested our proposed model with an external collected dataset and found a remarkable accuracy in that part too.

## 5. Result and Discussion

This section illustrates different evaluation methods for our experiments and discusses various aspects of the experiment. **Table 4** illustrates the comprehensive performance evaluation metrics, while **Figure 5** visualizes the accuracy curve, **Figure 6** shows the Receiver Operating Characteristics, and **Figure 7** illustrates the confusion matrix of the top 4 baseline models used in our experiments. The selection of baseline models was meticulously based on their performance, prioritizing those that exceeded a 95% accuracy threshold and demonstrating minimal loss rates. In particular, the initial three models surpassed the 99% accuracy mark; however, DenseNet201 notably exhibited the lowest validation loss among them. Also, **Table 5** represents classification report of our selected top-baseline models.

**Table 4.** Accuracy and Loss Values for the Top 4 Models

| Model No. | Model Name | Validation Accuracy | Validation Loss |
|---|---|---|---|
| 1 | EfficientNetV2L | 0.9924 | 0.0243 |
| 2 | VGG16 | 0.9913 | 0.0537 |
| 3 | MobileNetV2 | 0.9904 | 0.0427 |
| 4 | DenseNet201 | 0.9747 | 0.2095 |

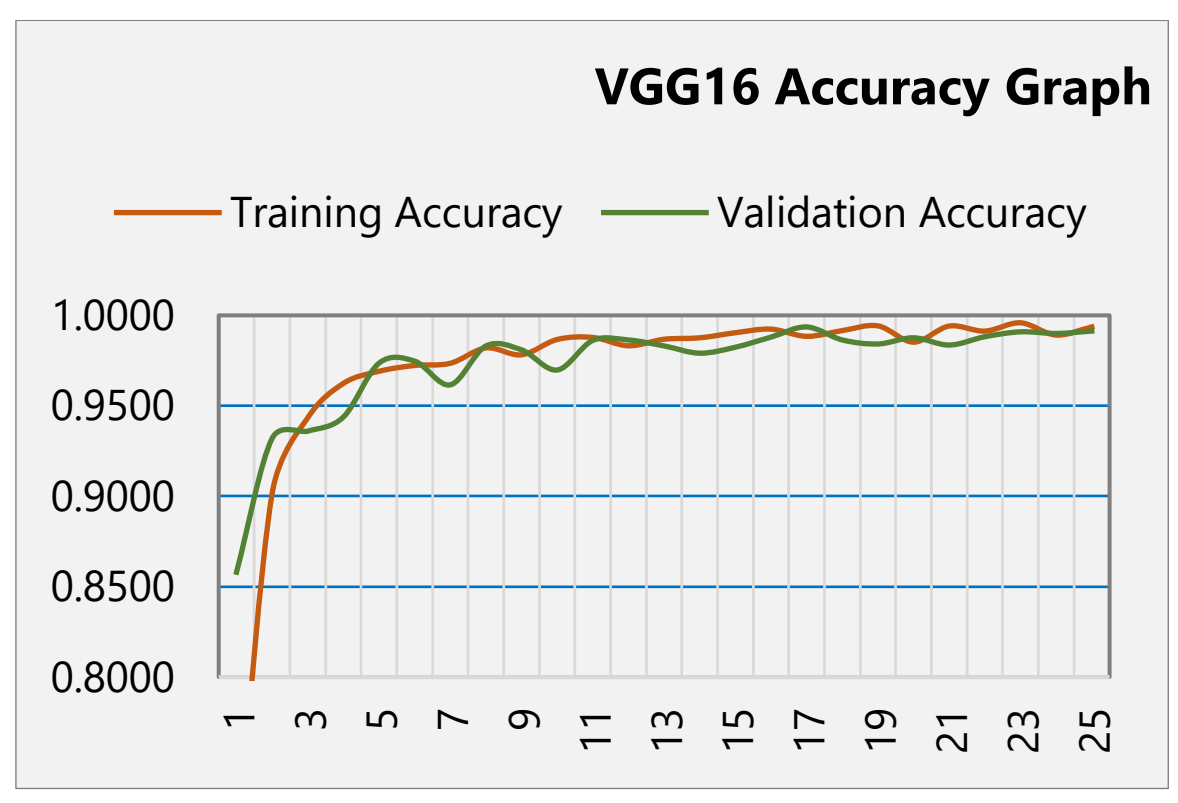

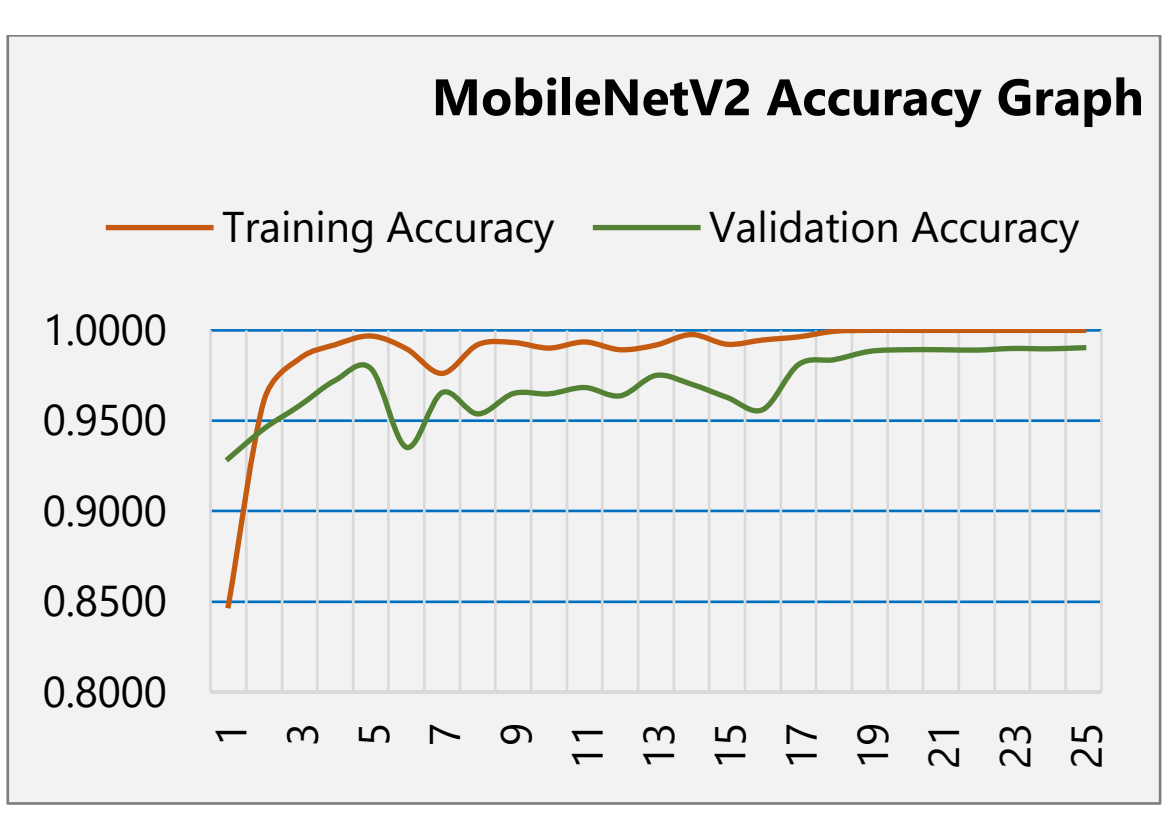

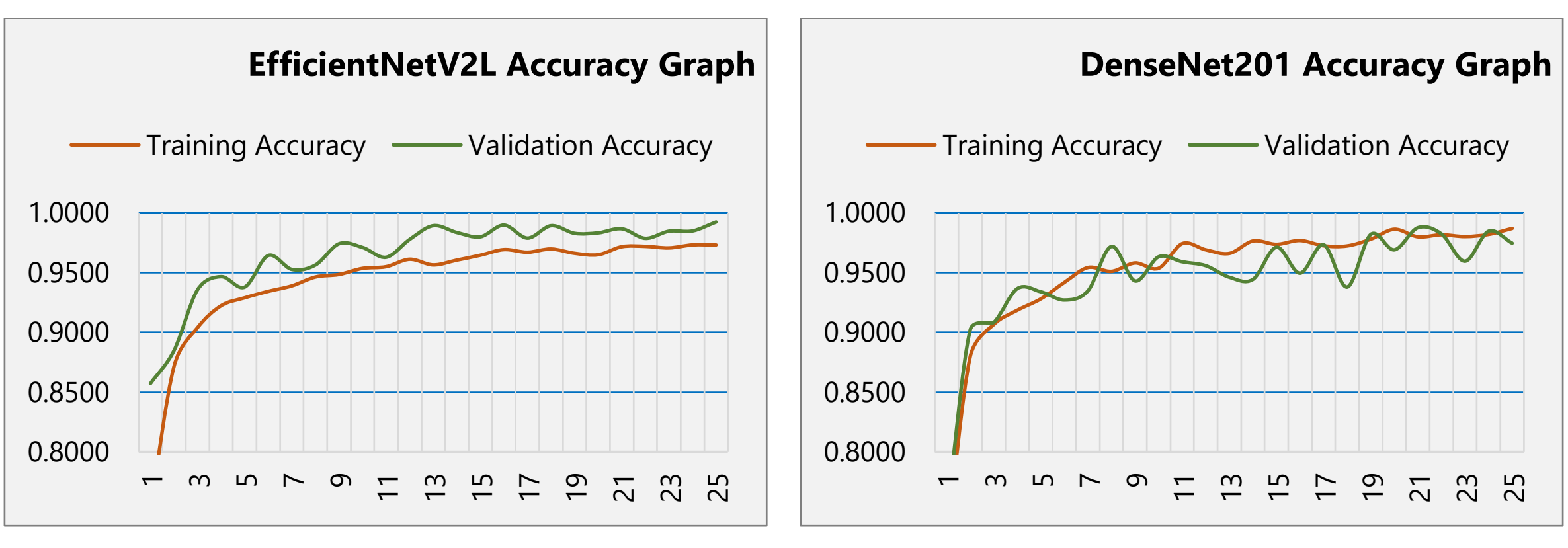

**Figure 5.** Training and Validation Accuracy Graph for the Top 4 Baseline Models

In the diagram, the values 0 to 19 represents Subol Lota, Bashmoti (Deshi), Ganjiya, Shampakatari, Sugandhi Katarivog, BR-28, BR-29, Paijam, Bashful, Lal Aush, BR-Jirashail, Gutisharna, Birui, Najirshail, Red Cargo, Polao (Katari), Polao (Chinigura), Amon, Shorna-5, and Lal Binni consecutively. The ROC values of EfficientNetV2L, VGG16, MobileNetV2, DenseNet201 are 0.99999945876, 0.99999925977, 0.9999985491, 0.99999825273 respectively.

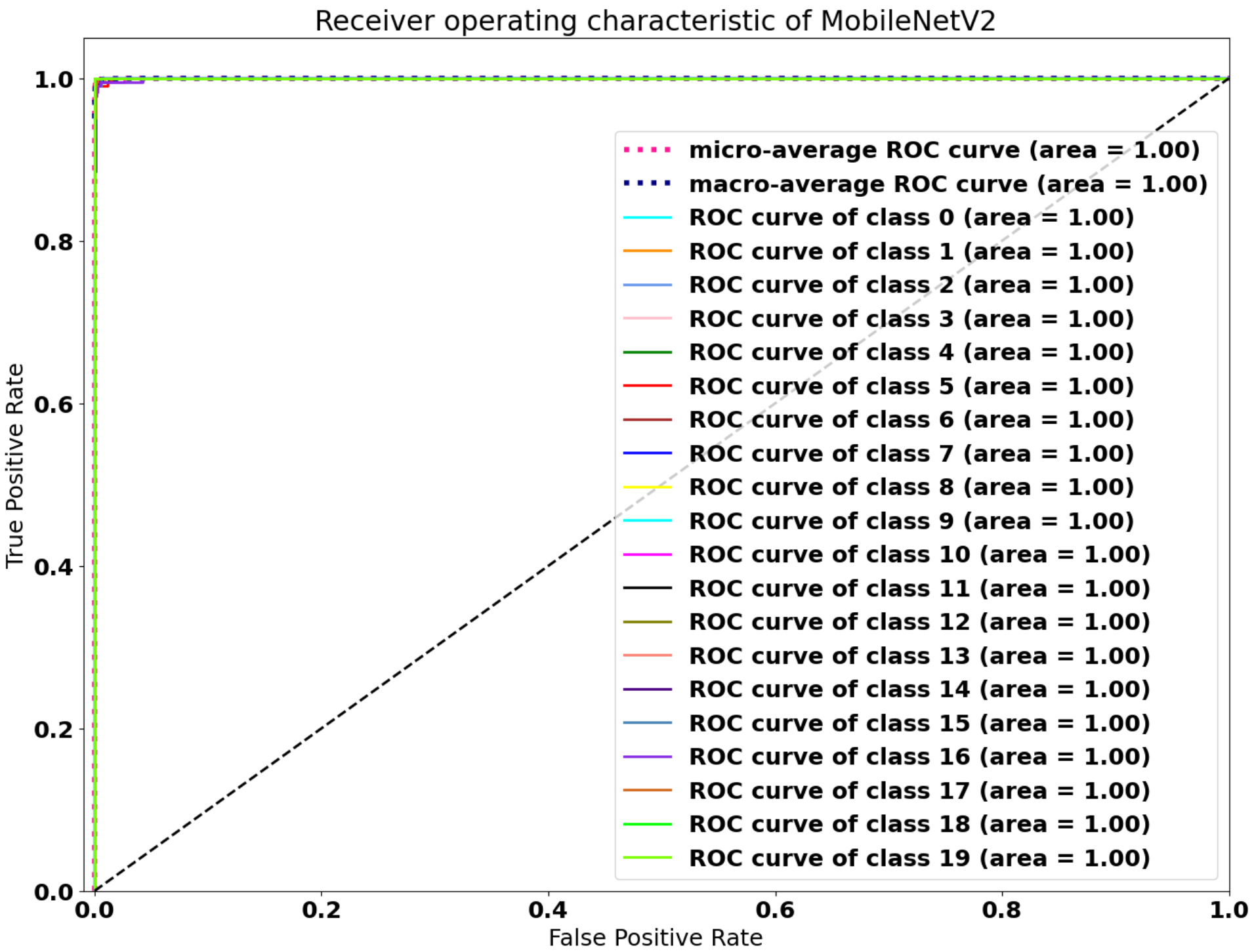

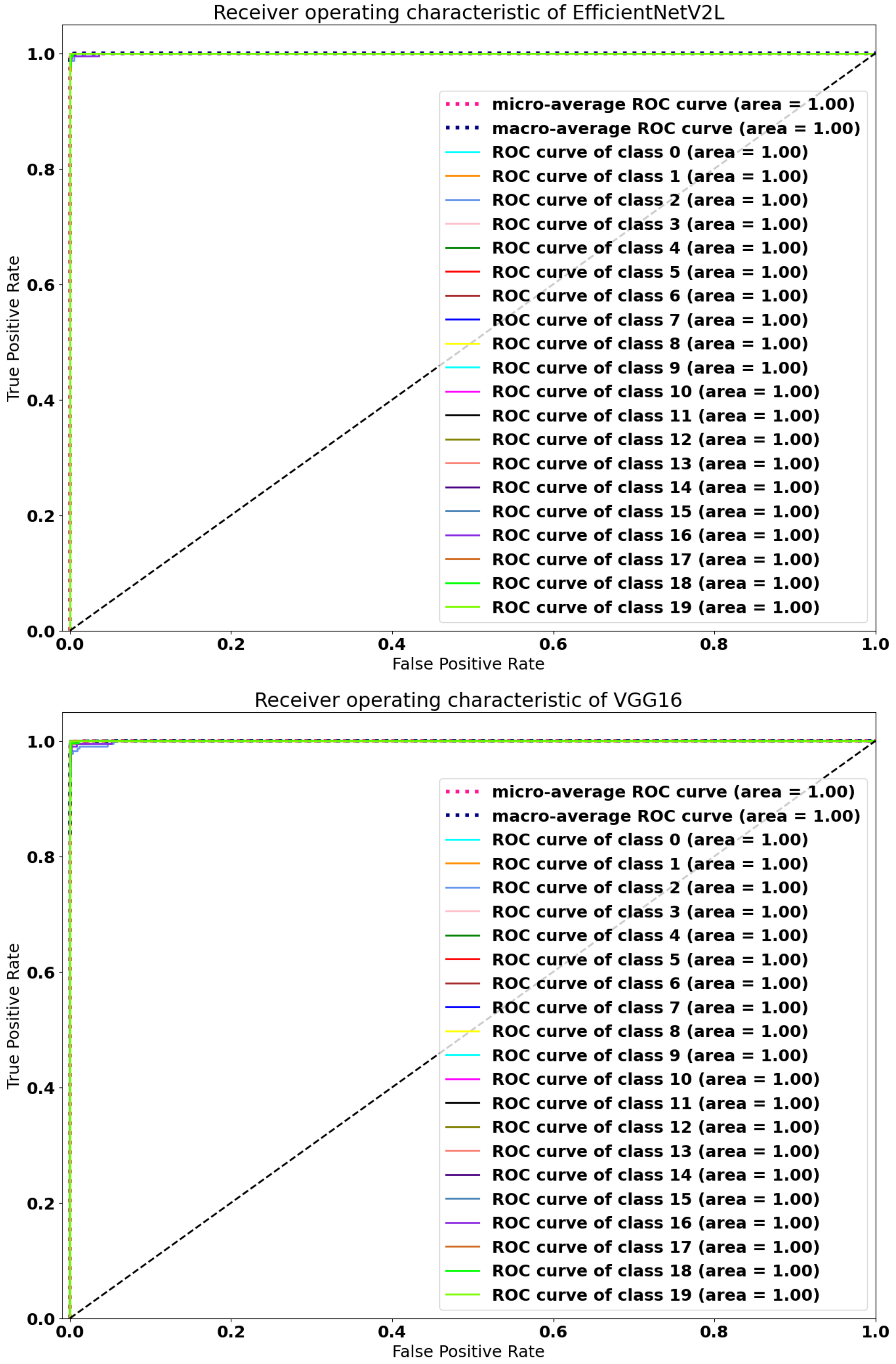
Receiver operating characteristic of EfficientNetV2L
True Positive Rate
False Positive Rate
micro-average ROC curve (area = 1.00)
macro-average ROC curve (area = 1.00)
ROC curve of class 0 (area = 1.00)
ROC curve of class 1 (area = 1.00)
ROC curve of class 2 (area = 1.00)
ROC curve of class 3 (area = 1.00)
ROC curve of class 4 (area = 1.00)
ROC curve of class 5 (area = 1.00)
ROC curve of class 6 (area = 1.00)
ROC curve of class 7 (area = 1.00)
ROC curve of class 8 (area = 1.00)
ROC curve of class 9 (area = 1.00)
ROC curve of class 10 (area = 1.00)
ROC curve of class 11 (area = 1.00)
ROC curve of class 12 (area = 1.00)
ROC curve of class 13 (area = 1.00)
ROC curve of class 14 (area = 1.00)
ROC curve of class 15 (area = 1.00)
ROC curve of class 16 (area = 1.00)
ROC curve of class 17 (area = 1.00)
ROC curve of class 18 (area = 1.00)
ROC curve of class 19 (area = 1.00)
Receiver operating characteristic of VGG16
True Positive Rate
False Positive Rate
micro-average ROC curve (area = 1.00)
macro-average ROC curve (area = 1.00)
ROC curve of class 0 (area = 1.00)
ROC curve of class 1 (area = 1.00)
ROC curve of class 2 (area = 1.00)
ROC curve of class 3 (area = 1.00)
ROC curve of class 4 (area = 1.00)
ROC curve of class 5 (area = 1.00)
ROC curve of class 6 (area = 1.00)
ROC curve of class 7 (area = 1.00)
ROC curve of class 8 (area = 1.00)
ROC curve of class 9 (area = 1.00)
ROC curve of class 10 (area = 1.00)
ROC curve of class 11 (area = 1.00)
ROC curve of class 12 (area = 1.00)
ROC curve of class 13 (area = 1.00)
ROC curve of class 14 (area = 1.00)
ROC curve of class 15 (area = 1.00)
ROC curve of class 16 (area = 1.00)
ROC curve of class 17 (area = 1.00)
ROC curve of class 18 (area = 1.00)
ROC curve of class 19 (area = 1.00)

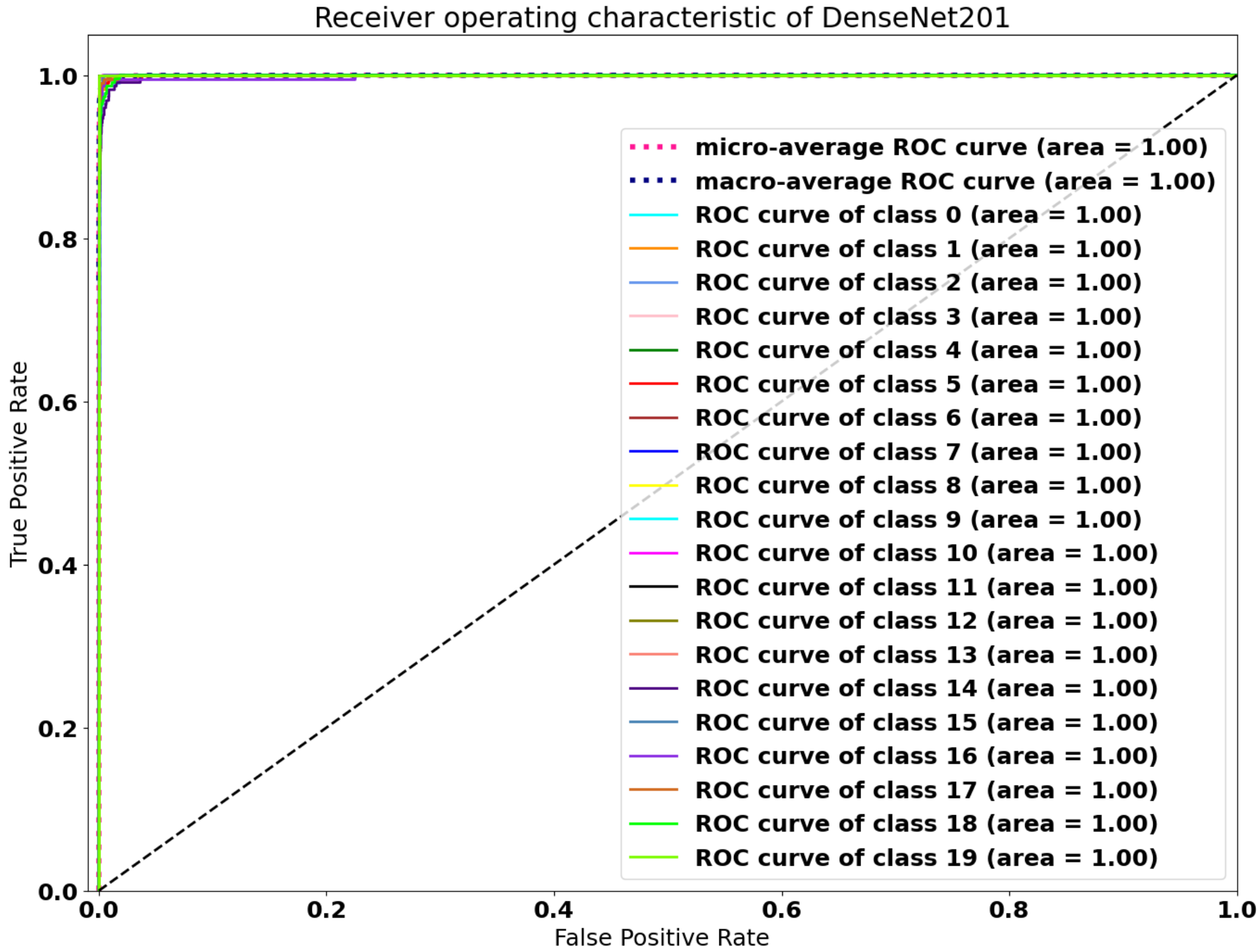

**Figure 6.** ROC for the Top 4 Baseline Models

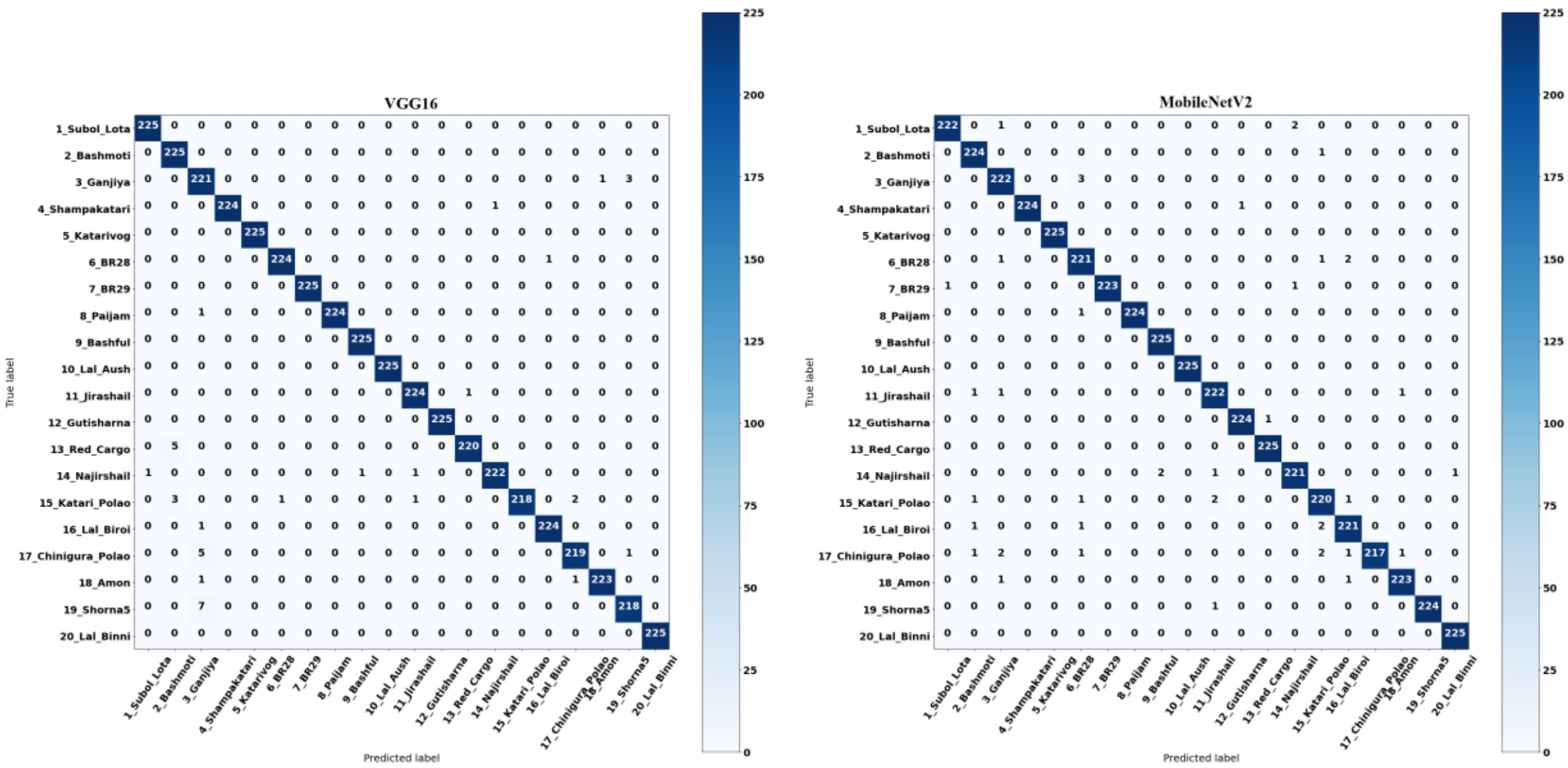

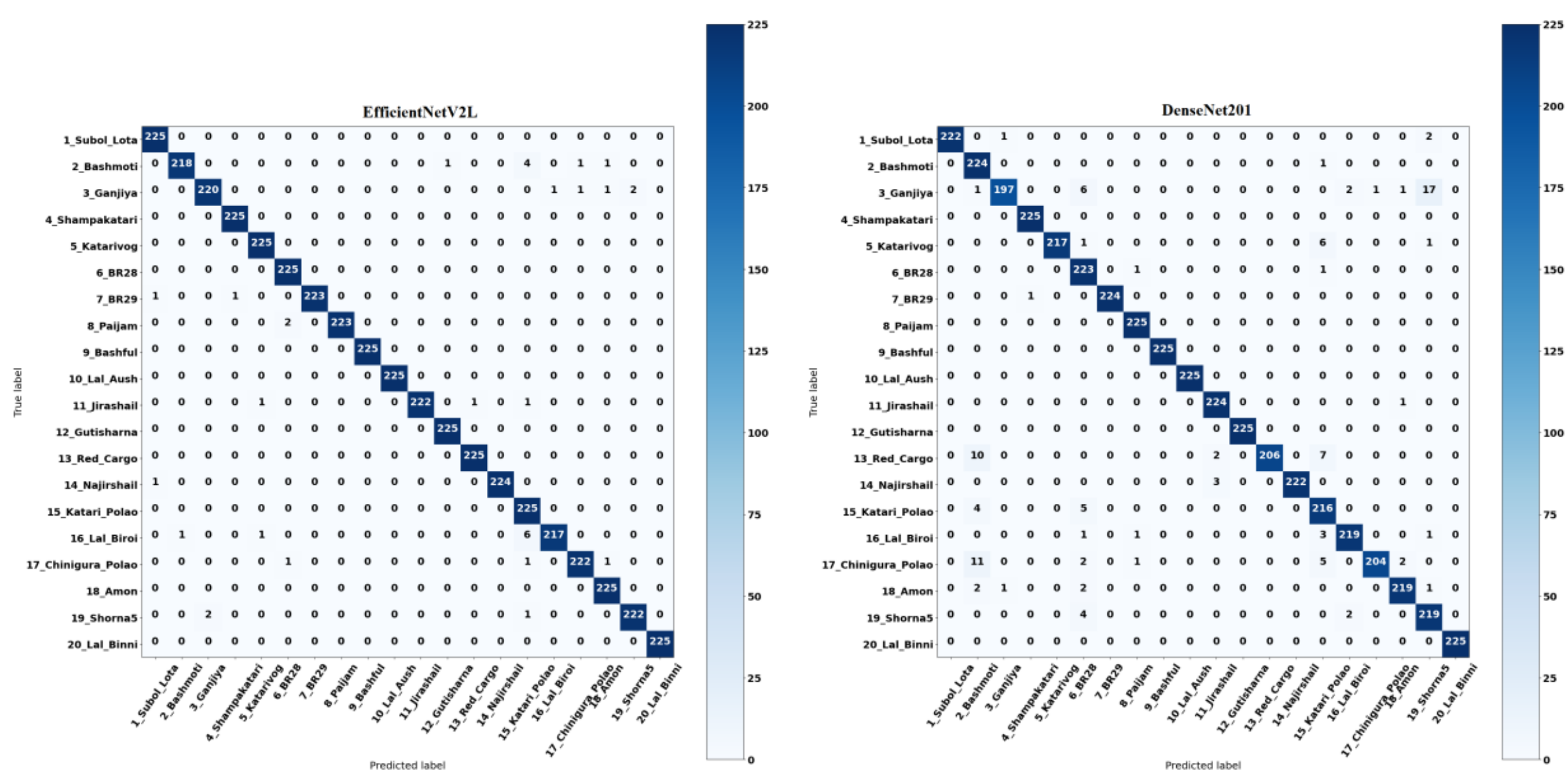

**Figure 7.** Confusion Matrix for the Top 4 Baseline Models

**Table 5.** Classification Report of Baseline Models

| Model | Class | Precision | Recall | f1-score | Model | Class | Precision | Recall | f1-score |
|---|---|---|---|---|---|---|---|---|---|
| EfficientNetV2L | 1_Subol_Lota | 0.99 | 1.00 | 1.00 | VGG16 | 1_Subol_Lota | 1.00 | 1.00 | 1.00 |
| | 2_Bashmoti | 1.00 | 0.97 | 0.98 | | 2_Bashmoti | 0.97 | 1.00 | 0.98 |
| | 3_Ganjiya | 0.99 | 0.98 | 0.98 | | 3_Ganjiya | 0.94 | 0.98 | 0.96 |
| | 4_Shampakatari | 1.00 | 1.00 | 1.00 | | 4_Shampakatari | 1.00 | 1.00 | 1.00 |
| | 5_Katarivog | 0.99 | 1.00 | 1.00 | | 5_Katarivog | 1.00 | 1.00 | 1.00 |
| | 6_BR28 | 0.99 | 1.00 | 0.99 | | 6_BR28 | 1.00 | 1.00 | 1.00 |
| | 7_BR29 | 1.00 | 0.99 | 1.00 | | 7_BR29 | 1.00 | 1.00 | 1.00 |
| | 8_Paijam | 1.00 | 0.99 | 1.00 | | 8_Paijam | 1.00 | 1.00 | 1.00 |
| | 9_Bashful | 1.00 | 1.00 | 1.00 | | 9_Bashful | 1.00 | 1.00 | 1.00 |
| | 10_Lal_Aush | 1.00 | 1.00 | 1.00 | | 10_Lal_Aush | 1.00 | 1.00 | 1.00 |
| | 11_Jirashail | 1.00 | 0.99 | 0.99 | | 11_Jirashail | 0.99 | 1.00 | 0.99 |
| | 12_Gutisharna | 1.00 | 1.00 | 1.00 | | 12_Gutisharna | 1.00 | 1.00 | 1.00 |
| | 13_Red_Cargo | 1.00 | 1.00 | 1.00 | | 13_Red_Cargo | 1.00 | 0.98 | 0.99 |
| | 14_Najirshail | 1.00 | 1.00 | 1.00 | | 14_Najirshail | 1.00 | 0.99 | 0.99 |
| | 15_Katari_Polao | 0.95 | 1.00 | 0.97 | | 15_Katari_Polao | 1.00 | 0.97 | 0.98 |
| | 16_Lal_Biroi | 1.00 | 0.96 | 0.98 | | 16_Lal_Biroi | 1.00 | 1.00 | 1.00 |
| | 17_Chinigura_Polao | 0.99 | 0.99 | 0.99 | | 17_Chinigura_Polao | 0.99 | 0.97 | 0.98 |
| | 18_Amon | 0.99 | 1.00 | 0.99 | | 18_Amon | 1.00 | 0.99 | 0.99 |
| | 19_Shorna5 | 0.99 | 0.99 | 0.99 | | 19_Shorna5 | 0.98 | 0.97 | 0.98 |
| | 20_Lal_Binni | 1.00 | 1.00 | 1.00 | | 20_Lal_Binni | 1.00 | 1.00 | 1.00 |

| Model | Class | Precision | Recall | f1-score | Model | Class | Precision | Recall | f1-score |
|---|---|---|---|---|---|---|---|---|---|
| MobileNetV2 | 1_Subol_Lota | 1.00 | 0.99 | 0.99 | DenseNet201 | 1_Subol_Lota | 1.00 | 0.99 | 0.99 |
| | 2_Bashmoti | 0.98 | 1.00 | 0.99 | | 2_Bashmoti | 0.89 | 1.00 | 0.94 |
| | 3_Ganjiya | 0.97 | 0.99 | 0.98 | | 3_Ganjiya | 0.99 | 0.88 | 0.93 |
| | 4_Shampakatari | 1.00 | 1.00 | 1.00 | | 4_Shampakatari | 1.00 | 1.00 | 1.00 |
| | 5_Katarivog | 1.00 | 1.00 | 1.00 | | 5_Katarivog | 1.00 | 0.96 | 0.98 |
| | 6_BR28 | 0.97 | 0.98 | 0.98 | | 6_BR28 | 0.91 | 0.99 | 0.95 |
| | 7_BR29 | 1.00 | 0.99 | 1.00 | | 7_BR29 | 1.00 | 1.00 | 1.00 |
|  | 8_Paijam | 1.00 | 1.00 | 1.00 |  | 8_Paijam | 0.99 | 1.00 | 0.99 |
|  | 9_Bashful | 0.99 | 1.00 | 1.00 |  | 9_Bashful | 1.00 | 1.00 | 1.00 |
|  | 10_Lal_Aush | 1.00 | 1.00 | 1.00 |  | 10_Lal_Aush | 1.00 | 1.00 | 1.00 |
|  | 11_Jirashail | 0.98 | 0.99 | 0.98 |  | 11_Jirashail | 0.98 | 1.00 | 0.99 |
|  | 12_Gutisharna | 1.00 | 1.00 | 1.00 |  | 12_Gutisharna | 1.00 | 1.00 | 1.00 |
|  | 13_Red_Cargo | 1.00 | 1.00 | 1.00 |  | 13_Red_Cargo | 1.00 | 0.92 | 0.96 |
|  | 14_Najirshail | 0.99 | 0.98 | 0.98 |  | 14_Najirshail | 1.00 | 0.99 | 0.99 |
|  | 15_Katari_Polao | 0.97 | 0.98 | 0.98 |  | 15_Katari_Polao | 0.90 | 0.96 | 0.93 |
|  | 16_Lal_Biroi | 0.98 | 0.98 | 0.98 |  | 16_Lal_Biroi | 0.98 | 0.97 | 0.98 |
|  | 17_Chinigura_Polao | 1.00 | 0.96 | 0.98 |  | 17_Chinigura_Polao | 1.00 | 0.91 | 0.95 |
|  | 18_Amon | 0.99 | 0.99 | 0.99 |  | 18_Amon | 0.98 | 0.97 | 0.98 |
|  | 19_Shorna5 | 1.00 | 1.00 | 1.00 |  | 19_Shorna5 | 0.91 | 0.97 | 0.94 |
|  | 20_Lal_Binni | 1.00 | 1.00 | 1.00 |  | 20_Lal_Binni | 1.00 | 1.00 | 1.00 |

In our experimental setup, we utilize ensemble learning by stacking the top 4 baseline models with 14 meta-learners. This rigorous experimentation leads to two key observations: First, the XGBoost meta-classifier consistently achieves 100% accuracy across all top model combinations, underscoring its robust performance and reliability. Second, all meta-learners demonstrate their effectiveness by delivering the highest accuracy for the top model combinations (1, 2, 3, and 4), which are, respectively, EfficientNetV2L, VGG16, MobileNetV2, and DenseNet201. This emphasizes the collective strength of the ensemble approach, wherein the meta-learners significantly enhance the predictive capabilities of the base models, resulting in superior performance across various combinations. **Table 6** provides detailed results for the different top model combinations.

**Table 6.** Accuracy comparison of the ensemble method using 14 meta-classifiers. (✦) indicates the top stacked models and selects the best metaclassifier

| **Model Combinations** | **Logistic Regression** | **K Nearest Neighbour** | **Support Vector Classifier** | **Decision Tree** | **Random Forest** | **AdaBoost Classifier** | **XGB Classifier✦** | **GB Classifier** | **Gaussian Naïve Bayes** | **Passive Aggressive Classifier** | **Ridge Classifier** | **Stochastics Gradient Descent Classifier** | **Extra Trees Classifier** | **MLP Classifier** |
|---|---|---|---|---|---|---|---|---|---|---|---|---|---|---|
| TOP (1,2) | 0.961 | 0.919 | 0.964 | 0.974 | 0.984 | 0.951 | 1.000 | 0.976 | 0.956 | 0.975 | 0.959 | 0.963 | 0.979 | 0.971 |
| TOP (1,2,3) | 0.985 | 0.949 | 0.989 | 1.000 | 1.000 | 0.969 | 1.000 | 0.995 | 0.986 | 0.980 | 0.985 | 0.988 | 1.000 | 0.989 |
| **TOP (1,2,3,4) ✦** | **0.985** | **0.961** | **0.991** | **1.000** | **1.000** | **0.955** | **1.000** | **0.995** | **0.985** | **0.985** | **0.985** | **0.987** | **1.000** | **0.991** |

To investigate the robustness of our proposed ensemble approach, we used cross-validation as a standard evaluation method. **Figure 8** show how the 5-fold cross-validation is conduct in this experiment. The dataset is randomly divided into five equal subsets (or "folds"). In each iteration, four folds are used for training the model, while the remaining one is used for validation. This process repeats five times, ensuring that each fold serves as the validation set once. The final model performance is determined by averaging the accuracy, precision, recall, and F1-score across all five iterations. **Figure 9** represents the Box-Whisker plot for our robustness experiments which visually represents the distribution of the model's performance metrics. Models such as MLP (Multi-Layer Perceptron), GB (Gradient Boosting), and XGB (Extreme Gradient Boosting) demonstrate high and consistent performance, as indicated by their compact boxes and small whiskers, which reflect minimal variability across folds. These models also achieve higher median values, showcasing their robustness and reliability. In contrast, models like KNN (K-Nearest Neighbors) and ADB (AdaBoost) exhibit wider boxes and longer whiskers, suggesting greater variability and sensitivity to the specific data split in each fold. This variability may point to challenges in generalization or dependence on specific features in the dataset.

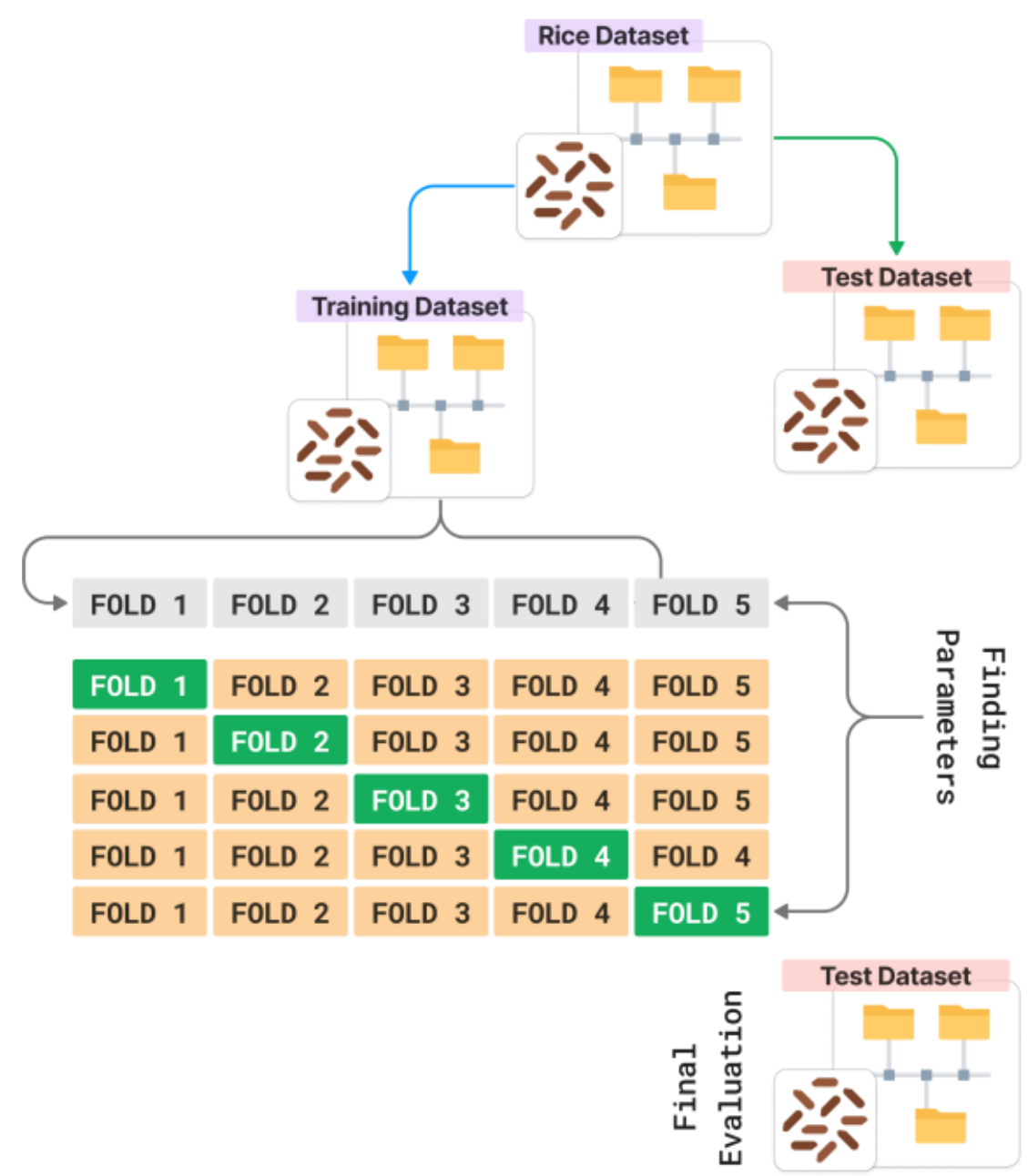

**Figure 8.** 5-Fold Cross-Validation Method

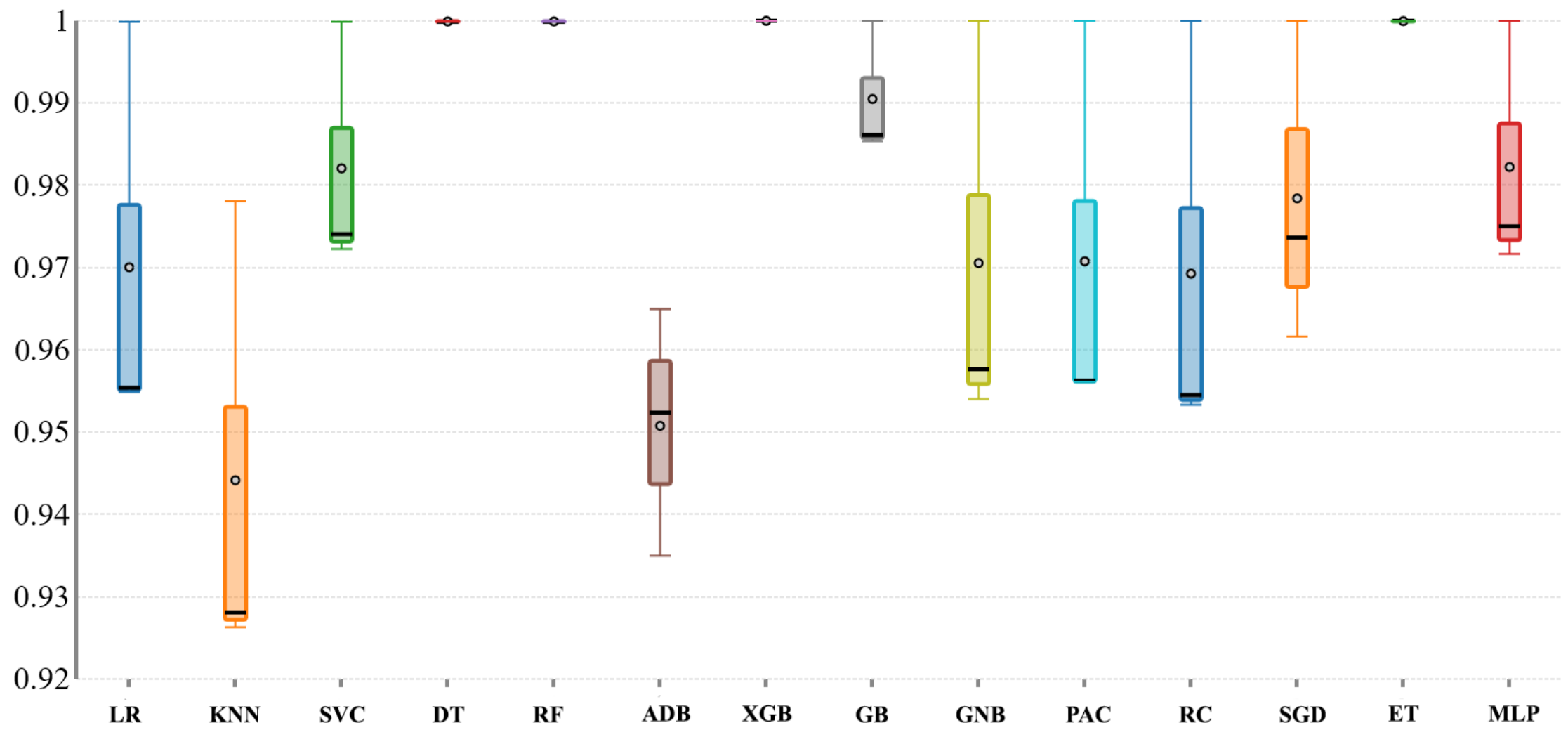

**Figure 9.** Cross-Validation Visualization using a Box-Whisker Plot

To evaluate the robustness and generalization capability of our proposed model, we tested it on an external dataset that was entirely unseen during training and validation. This external dataset was collected independently and possesses diverse characteristics, ensuring a realistic evaluation scenario. The confusion matrix in **Figure 10** demonstrates the classification performance of the proposed model on an external validation dataset, consisting of 700 samples per class across five rice varieties. The model achieved perfect classification for Arborio, Basmati, Ipsala, and Karacadag, with 100% accuracy in each category. For Jasmine, 698 samples were correctly classified, with only 2 misclassifications, resulting in an accuracy of 99.71% for this variety. These findings emphasize the model's potential for real-world applications, particularly in scenarios where accurate classification of rice varieties is critical. The minor misclassification in Jasmine highlights the need for further refinement in feature extraction, particularly for visually similar varieties. Overall, the confusion matrix underscores the high accuracy and generalization capability of the proposed ensemble-based classification approach. **Figure 11** represents some example test results of the external dataset for our model.

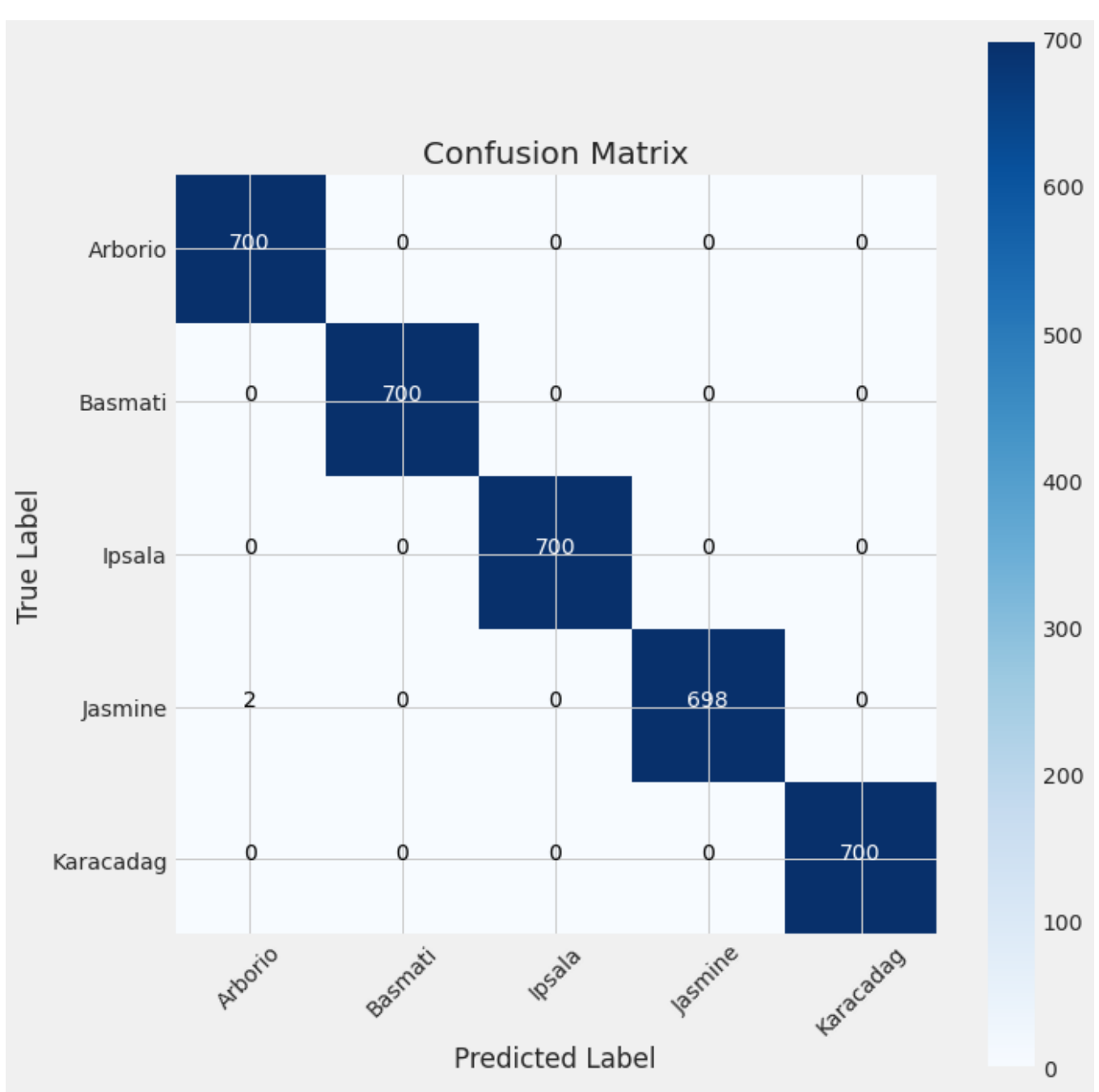

**Figure 10.** Confusion Matrix for Externa Data Validation

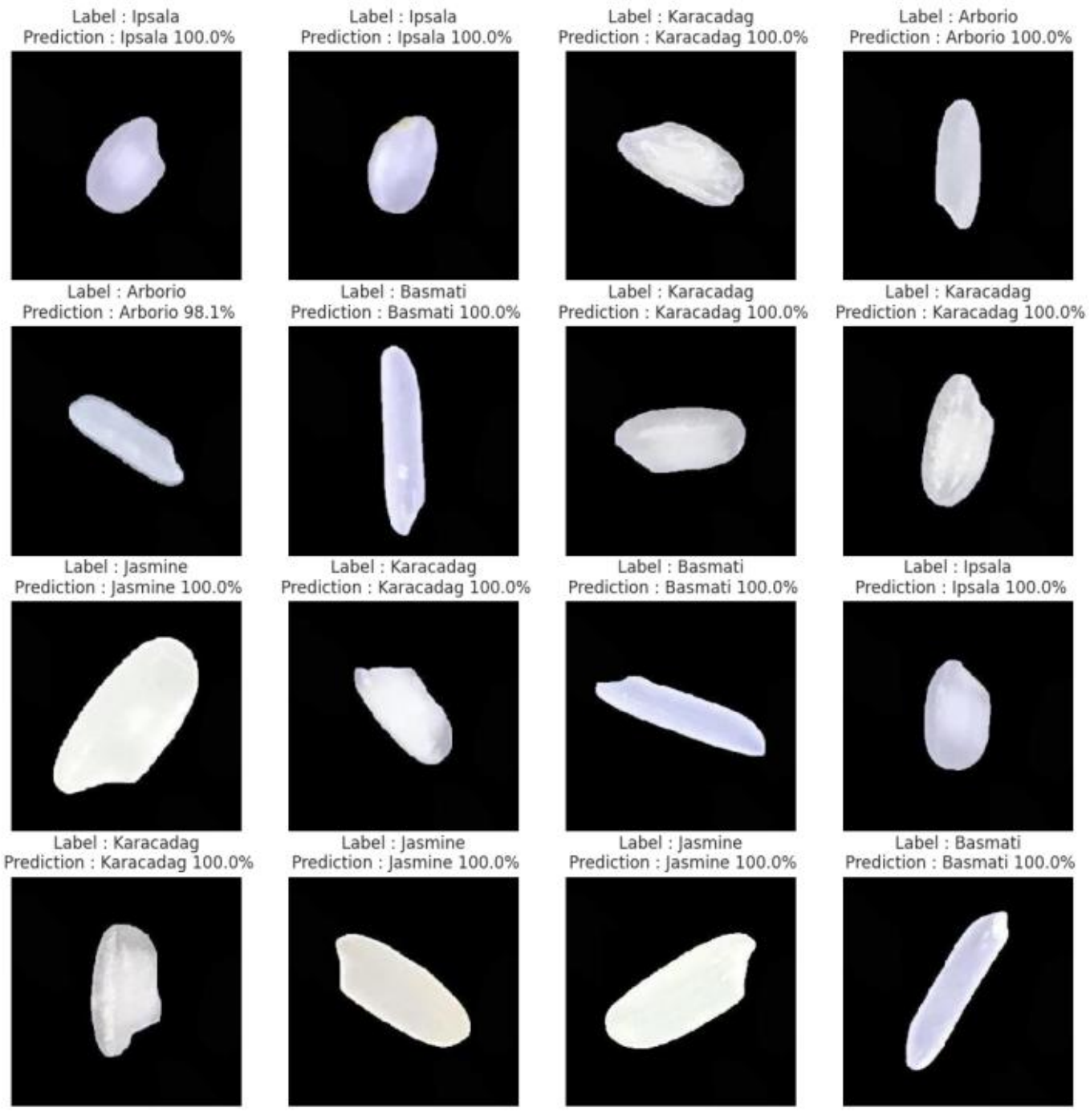

**Figure 11.** Example Test Result for Externa Data Validation

To prove the model’s usability, the model was converted into the TensorFlow Lite format to ensure compatibility and efficiency for deployment. This conversion process included model quantization to reduce size and improve inference speed, making it suitable for real-time usage on resource-constrained devices. Next, the optimized TensorFlow Lite model was deployed to a cloud server, enabling centralized access and scalability. A dedicated API endpoint was established to handle inference requests from client devices. This setup not only ensured secure and reliable interactions between the app and the model but also enabled quick updates and maintenance of the backend model without disrupting the user experience. To make this powerful model accessible to users, we developed a beta version Android application with an intuitive interface. The app allows users to capture images of rice grains using their smartphone cameras at 5x zoom or higher, ensuring the details necessary for accurate classification are preserved. Upon capturing an image, the app sends it to the cloud server via the API, where the deployed model processes the input and returns the predicted rice variety. The result is then displayed to the user in real-time. **Figure 12** illustrates the process of machine learning model deployment and **Figure 13** illustrates the app interfaces and test examples.

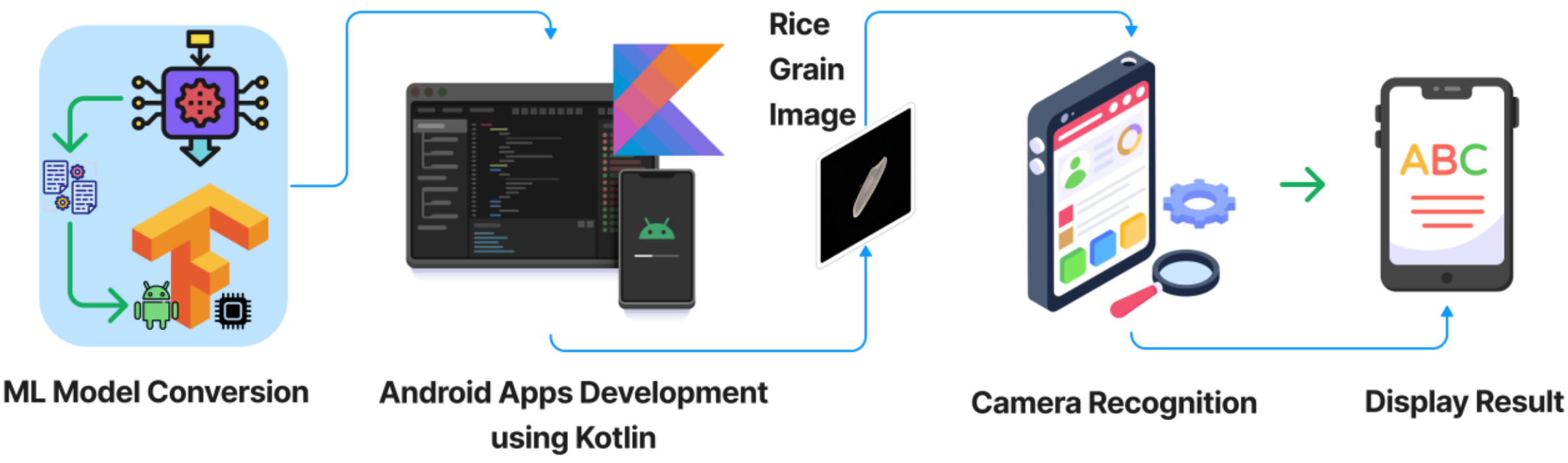

**Figure 12.** Model Deployment Process

The deployment of the model into a smartphone application introduces practical usability challenges, including variable lighting, camera quality, and user errors in capturing images. Preliminary testing showed minor performance degradation under non-ideal conditions. To address these issues, future iterations of the app will incorporate user guidance features and real-time image preprocessing techniques to enhance performance in diverse environments.

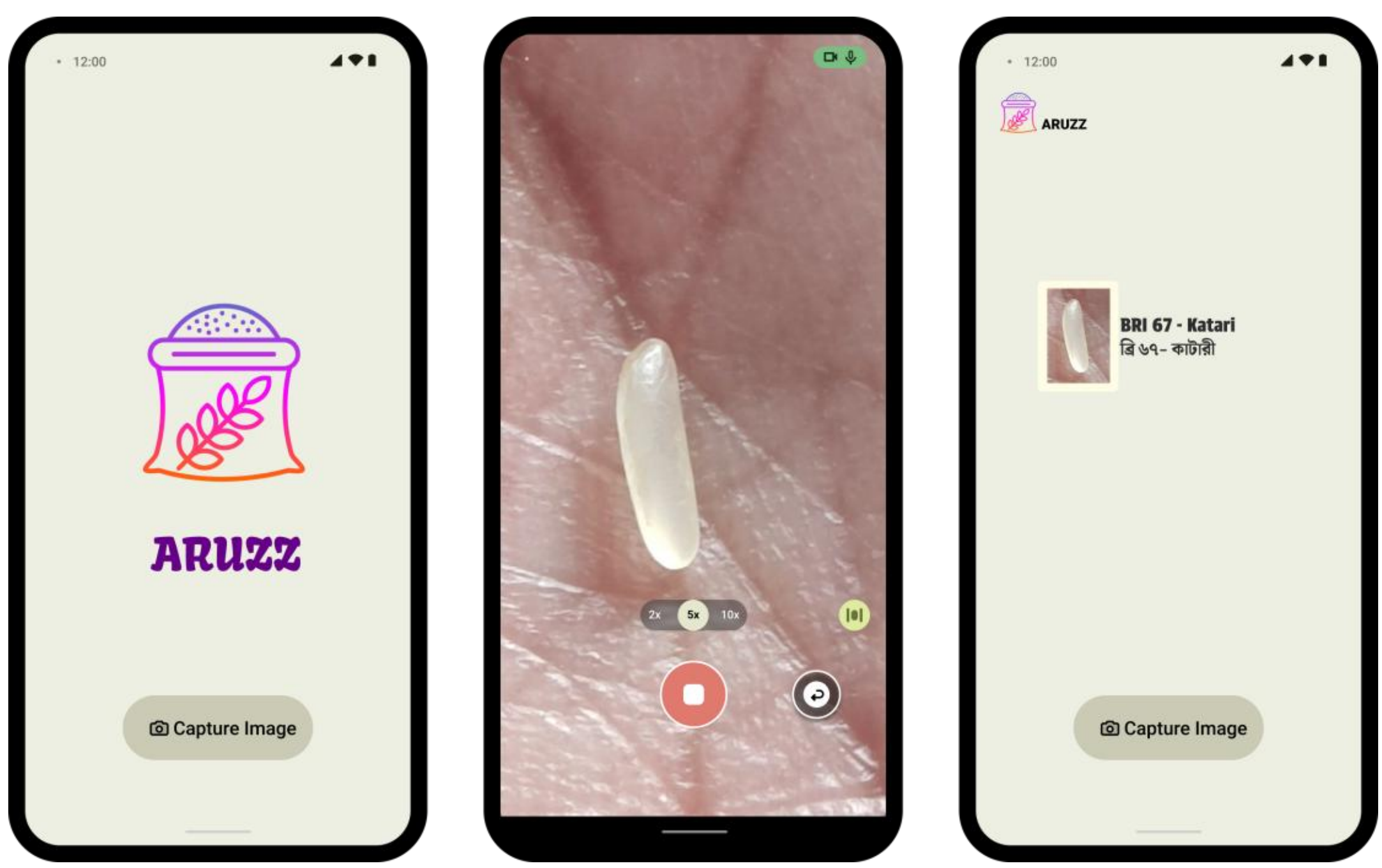

**Figure 13.** App Interface and Test Example

While our model achieved 100% accuracy on the self-curated dataset, we acknowledge potential limitations, including overfitting risks, dependency on high-quality images, and challenges in handling highly similar varieties in uncontrolled environments. Additionally, while our CNN-based ensemble model demonstrated superior performance, exploring Vision Transformers in future work may provide valuable insights into their applicability for rice variety classification.

## 6. Comparative Result Analysis

Since 1995, over 30 research experiments have been published exclusively on rice variety classification. Over the past three decades, this field has seen significant advancements, including an increase in the number of rice varieties studied, the availability of diverse datasets, and the adoption of advanced methodologies such as deep learning, machine learning, artificial neural networks, and fuzzy systems. Building on these developments, our proposed ensemble model has achieved unprecedented success, attaining the highest recorded performance levels. By leveraging a comprehensive dataset of 20 different rice varieties, the model has surpassed all prior benchmarks, marking a major milestone in rice variety classification research. **Table 7** presents a comparison of rice variety classification results over the past 10 years. Most of these studies utilized either self-curated datasets or publicly available popular datasets. From the summary comparison, it is evident that while some included a large number of rice varieties, their dataset sizes were small. Others had large datasets but lacked diversity in rice varieties. Most studies failed to achieve a balanced variation in dataset size, variety, and morphological as well as color-textural differences. In contrast, our study offers a comprehensive dataset featuring 20 rice varieties with diverse sizes, shapes, and color-texture variations, achieving the best classification accuracy and providing a real-life applicable app interface.

**Table 7.** Results from Comparison with Other Studies

| Year | Authors | Dataset | Method | Accuracy |
|---|---|---|---|---|
| 2015 | Huang et al. [13] | 79 varieties (316 data) | Support Vector Machine techniques | 79.74% |
| | Aki et al. [14] | 4 varieties (100 data) | Non-Nested Generalization Algorithm | 90.5% |
| | Sumaryanti et al. [15] | 10 varieties (1000 data) | Learning Vector Quantization Neural Network Algorithm | 96.6% |
| 2016 | Singh and Chaudhury [16] | 4 varieties | Back-propagation neural network | ~96% |
| | Zareiforoush et al. [18] | 4 varieties | Artificial Neural Network | 98.72% |
| 2017 | Lin et al. [19] | 3 varieties (3819 data) | Deep convolutional neural network | 95.5% |
| | Rexce J et al. [20] | 13 varieties | MLP Neural Network | 92.31% |
| 2018 | Wah et al. [21] | 3 varieties (329 samples) | Image segmentation and the K-Nearest Neighbor classifier | 83-100% |
| 2019 | Cinar and Koklu [22] | 2 varieties (3810 data) | Artificial intelligence methods | 93.02% |
| | Hue et al. [23] | (200 data) | Artificial neural networks | 95.5% |
| | Dheer and Singh [24] | 8 varieties (800 data) | Machine learning classifiers | 99.16% |
| | Kumar and Javeed [25] | 5 varieties (500 data) | SVM | 92% |
| 2020 | Ibrahim et al. [26] | 3 varieties | Artificial neural networks | 93.34% |
| | Weng et al. [27] | 10 varieties (4320 data) | Hyperspectral imaging and deep learning techniques | 98.57% |
| 2022 | Saxena et al. [28] | 5 varieties (75000 data) | Machine learning methods | 99.85% |
| | Lee and Tay [29] | 7 varieties (700 data) | Convolutional Neural Networks | 99.5% |
| | Nair et al. [30] | 2 varieties (3810 data) | A genetic fuzzy cascading system | 94% |
| | Jeyaraj et al. [31] | 5 varieties (750 data) | Deep learning technique | 98.2% |
| 2023 | Tasci et al. [32] | 5 varieties (75000 data) | LeeNet-5-based quantized neural network method | 99.87% |
| | Iqbal et al. [33] | 5 varieties (75000 data) | MobileNetV2 model | 99.7% |
| 2024 | Setiawan et al. [66] | 3 varieties (45000 data) | Ensemble Learning: SVM | 96.74% |
| | Kang et al. [67] | 5 varieties (550 datasets) | Principal Component Analysis, BP Neural Network, Random Forest | 95.30% |
| | **Proposed Method** | **20 varieties (27000 data)** | **Stacking-based Ensemble Method** | **100%** |

## 7. Conclusions and Future Directions

This article presents a comprehensive machine learning approach combined with a practical Android-based expert system for classifying rice varieties. The study utilized transfer learning models with hyperparameter tuning to extract deep features from rice grain images and employed a stacking-based ensemble method to enhance classification accuracy. This approach achieved an exceptional accuracy of 100%. For future work, several avenues can be explored to enhance the robustness and practicality of rice variety classification. First, expanding the dataset to include a larger and more diverse collection of rice varieties is essential. This expansion should involve sourcing rice samples from multiple geographical regions, capturing images in uncontrolled environments, and incorporating additional rice varieties. By doing so, the model will better reflect real-world conditions, such as mixed-grain samples, variations caused by harvesting processes, and environmental factors. Synthetic data augmentation techniques that simulate real-world noise and occlusions can also be employed to further improve the model's adaptability. To address computational efficiency, lightweight models such as MobileNet variants could be employed to accelerate classification on resource-constrained devices. Additionally, deploying the model in a mobile application presents challenges like variable lighting conditions, user inconsistencies, and hardware differences. To mitigate these issues, we propose real-time image preprocessing techniques, including adaptive brightness correction, contrast enhancement, and automated focus adjustment. Providing users with guidance on optimal image capture conditions—such as maintaining specific zoom levels and ensuring adequate lighting—can further enhance classification accuracy. In terms of methodological advancements, implementing Vision Transformers for rice variety classification could enable comparative analysis with CNN-based approaches, helping identify potential performance improvements and trade-offs. Refining feature extraction techniques, particularly for distinguishing highly similar rice varieties, will also play a critical role. Furthermore, validating the model on larger, more diverse datasets in uncontrolled environments will ensure its robustness and applicability to real-world scenarios. Another promising approach involves adopting a federated learning framework. This would allow researchers from different regions to locally train a global model with diverse rice image data, promoting better generalization across varied datasets. Future iterations could explore this collaborative strategy, enabling a more inclusive and adaptable classification process.

## Acknowledgment

We extend our heartfelt gratitude to Mrs. Sadia for her invaluable assistance and unwavering support throughout our research endeavors. Her dedication and support have greatly enriched our work, and we deeply appreciate her contributions to our article.

## Funding

The authors did not receive support from any organization for the submitted work.

## Conflicts of interest

The authors have no conflicts of interest to declare that are relevant to the content of this article.

## References

Bandumula, N. (2017). Rice production in Asia: Key to global food security. Krishi. 88(4).

Chauhan, B.S., Jabran, K., & Mahajan, G. (2017). Rice production worldwide. Springer International Publishing, Cham. https://doi.org/10.1007/978-3-319-47516-5

Çifci, A., & Kırbaş, İ. (2024). Fusion of machine learning and explainable AI for enhanced rice classification: A case study on Cammeo and Osmancik species. European Food Research and Technology. https://doi.org/10.1007/s00217-024-04614-9

FAO. (2023). Crops and livestock products [WWW Document]. Food and Agriculture Organization of the United Nations. Retrieved from https://www.fao.org/faostat/en/#data/QCL

Wu, J., Yu, H., Dai, H., Mei, W., Huang, X., Zhu, S., & Peng, M. (2012). Metabolite profiles of rice cultivars containing bacterial blight-resistant genes are distinctive from susceptible rice. Acta Biochimica et Biophysica Sinica, 44, 650–659. https://doi.org/10.1093/abbs/gms043

Komal, Sethi, G.K., & Bawa, R.K. (2023). Automatic rice variety identification system: State-of-the-art review, issues, challenges, and future directions. Multimedia Tools and Applications. https://doi.org/10.1007/s11042-023-14487-x

Marini, F., Zupan, J., & Magrì, A.L. (2004). On the use of counterpropagation artificial neural networks to characterize Italian rice varieties. Analytica Chimica Acta, 510, 231–240. https://doi.org/10.1016/j.aca.2004.01.009

Hobson, D.M., Carter, R.M., & Yan, Y. (2007). Characterisation and identification of rice grains through digital image analysis. Conference Proceedings. https://doi.org/10.1109/imtc.2007.379116

Guzman, J.D., Peralta, E.K., Nagatsuka, T., Ninomiya, S. (2008). Classification of Philippine rice grains using machine vision and artificial neural networks. World Conference on Agricultural Information and IT, 41–48.

OuYang, A.-G., Gao, R., Liu, Y., Sun, X., Pan, Y., & Dong, X. (2010). An automatic method for identifying different varieties of rice seeds using machine vision technology. 2010 Sixth International Conference on Natural Computation. https://doi.org/10.1109/icnc.2010.5583370

Nagendra, M.S., Selvaraju, P., Jerlin, R., Ganesamurthy, K., & Senthil, N. (2020). Identification and characterization of popular rice (Oryza sativa L.) varieties through chemical tests. Journal of Phytology, 82–85. https://doi.org/10.25081/jp.2020.v12.6513

Huang, C., Liu, L., Yang, W., Xiong, L., & Duan, L. (2016). Rapid identification of rice varieties by grain shape and yield-related features combined with multi-class SVM. IFIP Advances in Information and Communication Technology, 390–398. https://doi.org/10.1007/978-3-319-48357-3_38

Aki, O., Güllü, A., & Uçar, E. (2016). Classification of rice grains using image processing and machine learning techniques.

Sumaryanti, L., Musdholifah, A., & Hartati, S. (2015). Digital image-based identification of rice variety using image processing and neural network. TELKOMNIKA Indonesian Journal of Electrical Engineering, 16, 182. https://doi.org/10.11591/tijee.v16i1.1602

Singh, Ksh.R., & Chaudhury, S. (2016). Efficient technique for rice grain classification using back-propagation neural network and wavelet decomposition. IET Computer Vision, 10, 780–787. https://doi.org/10.1049/iet-cvi.2015.0486

Kuo, T.-Y., Chung, C.-L., Chen, S.-Y., Lin, H.-A., & Kuo, Y.-F. (2016). Identifying rice grains using image analysis and sparse-representation-based classification. Computers and Electronics in Agriculture, 127, 716–725. https://doi.org/10.1016/j.compag.2016.07.020

Zareiforoush, H., Minaei, S., Alizadeh, M., & Banakar, A. (2016). Qualitative classification of milled rice grains using computer vision and metaheuristic techniques. Journal of Food Science and Technology, 53, 118–131. https://doi.org/10.1007/s13197-015-1947-4

Lin, P., Li, X., Chen, Y., & He, Y.-L. (2018). A deep convolutional neural network architecture for boosting image discrimination accuracy of rice species. Food and Bioprocess Technology, 11, 765–773. https://doi.org/10.1007/s11947-017-2050-9

Kingsly, U. (2015). Classification of milled rice using image processing. International Journal of Scientific & Engineering Research.

Wah, T.N., San, P.E., & Hlaing, T. (2018). Analysis of feature extraction and classification of rice kernels for Myanmar rice using image processing techniques. IJSRP. Retrieved from https://www.ijsrp.org/research-paper-0818.php?rp=P807680 (accessed 3.31.24)

Cinar, I. (2019). Classification of rice varieties using artificial intelligence methods. International Journal of Intelligent Systems and Applications in Engineering, 7, 188–194. https://doi.org/10.18201/ijisae.2019355381

Hu, Y., Du, Y., San, L., & Tian, J. (2019). Research on rice grain shape detection method based on machine vision. 2019 5th International Conference on Control, Automation and Robotics (ICCAR). https://doi.org/10.1109/iccar.2019.8813449

Dheer, P., & Singh, R.K. (2017). Identification of Indian rice varieties using machine learning classifiers. Plant Archives, 19(1), 155–158.

Kumar, M. Senthil, & Javeed, Md. (2019). An efficient rice variety identification scheme using shape, Harlick & color feature extraction and multiclass SVM. International Journal of Engineering and Advanced Technology, 8, 3629–3632. https://doi.org/10.35940/ijeat.f9362.088619

Ibrahim, S., Kamaruddin, S.B.A., Zabidi, A., & Md. Ghani, N.A. (2020). Contrastive analysis of rice grain classification techniques: multi-class support vector machine vs artificial neural network. IAES International Journal of Artificial Intelligence (IJ-AI), 9, 616. https://doi.org/10.11591/ijai.v9.i4.pp616-622

Weng, S., Tang, P., Yuan, H., Guo, B., Yu, S., Huang, L., & Xu, C. (2020). Hyperspectral imaging for accurate determination of rice variety using a deep learning network with multi-feature fusion. Spectrochimica Acta Part A: Molecular and Biomolecular Spectroscopy, 234, 118237. https://doi.org/10.1016/j.saa.2020.118237

Saxena, P., Priya, K., Goel, S., Aggarwal, P.K., Sinha, A., & Jain, P. (2022). Rice varieties classification using machine learning algorithms. Journal of Pharmaceutical Negative Results, 3762–3772. https://doi.org/10.47750/pnr.2022.13.S07.479

Lee, & Tay. (2022). Rice grain classification using convolution neural network with small dataset. 2022 Joint 12th International Conference on Soft Computing and Intelligent Systems and 23rd International Symposium on Advanced Intelligent Systems (SCIS&ISIS). https://doi.org/10.1109/scisisis55246.2022.10002003

Jeyaraj, A., Rajan, E. (2022). Computer-assisted real-time rice variety learning using deep learning network. Rice Science, 29, 489–498. https://doi.org/10.1016/j.rsci.2022.02.003

Tasci, A., Istanbullu, A., Kosunalp, S., Iliev, T., Stoyanov, I., & Beloev, I. (2023). An efficient classification of rice variety with quantized neural networks. Electronics, 12, 2285–2285. https://doi.org/10.3390/electronics12102285

Iqbal, M.J., Aasem, M., Ahmad, I., Alassafi, M.O., Bakhsh, S.T., Noreen, N., & Alhomoud, A. (2023). On application of lightweight models for rice variety classification and their potential in edge computing. Foods, 12, 3993. https://doi.org/10.3390/foods12213993

Islam, et al. (2024). A visual dataset for recognition of rice varieties. Data in Brief, 110442–110442. https://doi.org/10.1016/j.dib.2024.110442

Weerts, H.J.P., Mueller, A.C., & Vanschoren, J. (2020). Importance of tuning hyperparameters of machine learning algorithms. arXiv:2007.07588 [cs, stat].

Galar, M., Fernández, A., Barrenechea, E., Bustince, H., & Herrera, F. (2011). An overview of ensemble methods for binary classifiers in multi-class problems: Experimental study on one-vs-one and one-vs-all schemes. Pattern Recognition, 44, 1761–1776. https://doi.org/10.1016/j.patcog.2011.01.017

Setiawan, R., & Oumarou, H. (2024). Classification of rice grain varieties using ensemble learning and image analysis techniques. Indonesian Journal of Data and Science, 5, 54–63. https://doi.org/10.56705/ijodas.v5i1.129

Kang, Z., Fan, R., Zhan, C., Wu, Y., Lin, Y., Li, K., Qing, R., & Xu, L. (2024). The rapid non-destructive differentiation of different varieties of rice by fluorescence hyperspectral technology combined with machine learning. Molecules, 29, 682–682. https://doi.org/10.3390/molecules29030682

Koklu, M., Cinar, I., & Taspinar, Y.S. (2021). Classification of rice varieties with deep learning methods. Computers and Electronics in Agriculture, 187, 106285. https://doi.org/10.1016/j.compag.2021.106285
Islam, M.M., Himel, G.M.S., Moazzam, M.G., & Uddin, M.S. (2024). Aruzz22.5K: An image dataset of rice varieties. Mendeley Data, V4, https://doi.org/10.17632/3mn9843tz2.4
Mousavi, J., Akhlaghian Tab, F., & Mollazade, K. (2011). Classification of rice varieties using optimal color and texture features and BP neural networks. 2011 7th Iranian Conference on Machine Vision and Image Processing. https://doi.org/10.1109/iranianmvip.2011.6121583
MousaviRad, S.J., Akhlaghian Tab, F., & Mollazade, K. (2012). Application of Imperialist Competitive Algorithm for feature selection: A case study on bulk rice classification. International Journal of Computer Applications, 40, 41–48. https://doi.org/10.5120/5068-7485
Mousavi, J., Akhlaghian Tab, F., & Mollazade, K. (2012). Design of an expert system for rice kernel identification using optimal morphological features and back propagation neural network. International Journal of Applied Information Systems, 3, 33–37.
Kaur, H., & Singh, B.S. (2024). Classification and grading rice using multi-class SVM. IJCEM International Journal of Computational Engineering & Management, 27, 75–80.
Kambo, R. (2014). Classification of Basmati rice grain variety using image processing and principal component analysis. International Journal of Computer Science and Technology (IJCTT), 11(4), 117. Retrieved from https://www.ijcttjournal.org/archives/ijctt-v11p117 (accessed 10.26.24).
Qadri, S., Aslam, T., Nawaz, S.A., Saher, N., Razzaq, A., Ur Rehman, M., Ahmad, N., Shahzad, F., & Furqan Qadri, S. (2021). Machine vision approach for classification of rice varieties using texture features. International Journal of Food Properties, 24, 1615–1630. https://doi.org/10.1080/10942912.2021.1986523
Kaur, S., & Singh, D. (2015). Geometric feature extraction of selected rice grains using image processing techniques. International Journal of Computer Applications, 124(5), 41–46.
Bhat, S.P., Sreedath Panat, & Arunachalam, N. (2017). Classification of rice grain varieties arranged in scattered and heap fashion using image processing. Proceedings of SPIE. https://doi.org/10.1117/12.2268802
Koklu, M., Cinar, I., & Taspinar, Y.S. (2021). Classification of rice varieties with deep learning methods. Computers and Electronics in Agriculture, 187, 106285. https://doi.org/10.1016/j.compag.2021.106285
Çinar, İ., & Köklü, M. (2021). Determination of effective and specific physical features of rice varieties by computer vision in exterior quality inspection. Selcuk Journal of Agriculture and Food Sciences, 35, 229–243. https://doi.org/10.15316/sjafs.2021.252